\documentclass[twoside]{article}
\usepackage{ecj,palatino,epsfig,latexsym,natbib}
\usepackage{enumitem}
\usepackage{amsmath}
\usepackage{amssymb}
\usepackage{color}
\usepackage{graphicx}
\usepackage{epstopdf}
\usepackage{algorithm,algpseudocode,algorithmicx}
\usepackage{amsmath,latexsym,amssymb,amsthm,array,amsfonts,algorithm,algpseudocode,booktabs,multirow,cuted,stfloats}
\usepackage{footmisc}
\usepackage{units}
\usepackage{tabularx}
\usepackage{multirow}
\usepackage{soul}

\usepackage{graphicx}
\usepackage[caption=false,font=footnotesize]{subfig}
\usepackage{float}

\begin{document}

\ecjHeader{x}{x}{xxx-xxx}{201X}{TC-SAEA for MOPs with Non-uniform Evaluation Times}{Xilu Wang, Yaochu Jin, Sebastian Schmitt, Markus Olhofer}
\title{\bf Transfer Learning Based Co-surrogate Assisted Evolutionary Bi-objective Optimization for Objectives with Non-uniform Evaluation Times}

\author{\name{\bf Xilu Wang} \hfill \addr{xilu.wang@surrey.ac.uk}\\
        \addr{Department of Computer Science, University of Surrey, Guildford, GU2 7XH, United Kingdom}
\AND
       \name{\bf Yaochu Jin} \hfill \addr{yaochu.jin@surrey.ac.uk}\\
        \addr{Department of Computer Science, University of Surrey, Guildford, GU2 7XH, United Kingdom}
\AND
       \name{\bf Sebastian Schmitt} \hfill \addr{sebastian.schmitt@honda-ri.de}\\
        \addr{Honda Research Institute Europe GmbH, Carl-Legien-Strasse 30, D-63073 Offenbach/Main, Germany}
\AND
       \name{\bf Markus Olhofer} \hfill \addr{markus.olhofer@honda-ri.de}\\
        \addr{Honda Research Institute Europe GmbH, Carl-Legien-Strasse 30, D-63073 Offenbach/Main, Germany}
}

\maketitle

\begin{abstract}

Most existing multiobjetive evolutionary algorithms (MOEAs) implicitly assume that each objective function can be evaluated within the same period of time. Typically. this is untenable in many real-world optimization scenarios where evaluation of different objectives involves different computer simulations or physical experiments with distinct time complexity. To address this issue, a transfer learning scheme based on surrogate-assisted evolutionary algorithms (SAEAs) is proposed, in which a co-surrogate is adopted to model the functional relationship between the fast and slow objective functions and a transferable instance selection method is introduced to acquire useful knowledge from the search process of the fast objective. Our experimental results on DTLZ and UF test suites demonstrate that the proposed algorithm is competitive for solving bi-objective optimization where objectives have non-uniform evaluation times.

\end{abstract}


\section{Introduction}
Many real-world applications can be formulated as multi-objective optimization problems (MOPs) that simultaneously optimize two or more objective functions  \citep{abraham2005evolutionary}. Usually, different objectives are conflicting to each other and there exists a set of optimal compromise solutions, known as Pareto optimal solutions. The whole set of Pareto optimal solutions in the decision space is called the Pareto set (PS), and the projection of PS in the objective space is called Pareto front (PF). Over the past decades, multi-objective evolutionary algorithms (MOEAs) have been very successful in solving MOPs as they can obtain a set of optimal solutions in a single run \citep{zhou2011multiobjective}. However, MOEAs require an excessive number of function evaluations (FEs) before achieving a set of acceptable non-dominated solutions, preventing them from being applied to MOPs where FEs involve computationally intensive simulations or costly physical experiments \citep{jin2002gecco}.  


The limited budget of FEs is one main challenge in using MOEAs to solve expensive MOPs and surrogate models are useful tools for overcoming the computational obstacle \citep{jin2011surrogate,allmendinger2017survey,jin2019data}. Specifically, surrogate-assisted evolutionary algorithms (SAEAs) construct cheap surrogate models to evaluate candidate solutions instead of calling the original expensive objective functions. Various efficient models have been adopted in SAEAs, including radial basis function networks \citep{sun2017surrogate,li2020surrogate}, polynomial regressions \citep{zhou2005study}, support vector machines \citep{bourinet2016rare}, random forest \citep{wang2020forest}, and Gaussian processes (GPs) \citep{chugh2018surrogate,zhou2005study}. As it is impossible to build very accurate and highest-quality surrogate models for realistic objective functions using only a limited amount of training data, it makes sense to use surrogate models and the original objective functions together to secure the convergence of SAEAs, which is termed model management \citep{jin2011surrogate}. More precisely, model management aims to identify a limited number of new candidate solutions to be evaluated by the original objective functions, which are then used for updating the surrogates. Along this line of research, Bayesian optimization has become very popular. Bayesian optimization refers to a class of black-box optimization algorithms that use GPs as surrogates in combination with an acquisition function (AF) which proposes new data samples to be evaluated with the true objective functions for model management \citep{shahriari2015taking}.

In most existing MOEAs and SAEAs, all objective functions are assumed to have similar computational complexity and consequently the evaluation of each objective takes similar amount of time. However, it is common that objective functions in MOPs have distinct complexities. This class of MOPs for objectives with non-uniform evaluation times (or latencies) is first introduced by \cite{allmendinger2013hang}, and a general problem definition is further derived by \cite{allmendinger2015multiobjective}. Based on their work, MOPs for objectives with non-uniform latencies corresponds to a situation where an MOP involves both the fast (computationally cheap) and slow (computationally expensive) objective functions or an MOP involves computationally intensive objectives where evaluation time of each objective drastically varies. For example, during the design optimization of an aircraft hull, crashworthiness assessment simulations take on the order of ten to a couple of hundred hours to complete, while the aerodynamic properties can be evaluated within few hours \citep{wang2007review}. In such cases, there is no reason to believe all objective functions in an MOP have similar computational complexity. In this work, we consider bi-objective optimization problems with a fast and a slow objective functions.

In the following, we introduce the related notations and assumptions of the problems under consideration.
\begin{itemize}
\item The slow (or delayed/expensive) objective function is denoted as $f_{s}$, while the fast (or non-delayed/cheap) one is denoted as $f_{f}$, and the corresponding Gaussian process models (GPs) are denoted as $GP_{s}$ and $GP_{f}$, respectively. The ratio of the evaluation time of $f_{s}$ to that of $f_{f}$ is denoted by $\tau$. Here, we assume that $\tau$ is an integer larger than 1. Solutions that are evaluated by both $f_{f}$ and $f_{s}$ are denoted as $\mathbf{X}$. Consequently, the slow and fast objectives of $\mathbf{X}$ are denoted by $\mathbf{Y}_{s}$ and $\mathbf{Y}_{f}$, respectively. Due to the big difference in the evaluation time of the slow and fast objectives, to make full use of the time during which the slow objective is being evaluated, additional solutions, denoted by $\mathbf{X}^{a}$ can be evaluated for the fast objective function ($f_f$), which are denoted by $\mathbf{Y}_{f}^{a}$. Here, we assume that evaluation of the fast and slow objective functions can be done in parallel.
\item It is assumed that the computational time for building surrogates and for implementing the genetic operators is negligible compared to that for evaluating the expensive objectives. Consequently, the total computational time available for solving the problem under consideration is defined by the total budget of function evaluations. Specifically, the total budget is defined as the maximum number of function evaluations for the slow objective $f_{s}$, donated as $FE_s^{max}$. Consequently, the maximum budget for the fast objective $FE_{f}^{max}$ equals $FE_{s}^{max}*\tau$.
\end{itemize}

Given the above assumptions, most existing MOEAs and SAEAs are not able to efficiently solve MOPs with different evaluation latencies because they need to wait for completing the evaluation of the slow objective denoted as $f_{s}$ \citep{allmendinger2013hang}. Only recently has some work been dedicated to dealing with such problems that considers objectives with non-uniform evaluation times. A pioneering study considering MOPs with delayed objectives has been done by \cite{allmendinger2013hang}. They investigated several strategies inspired by fitness inheritance to estimate the objective values (called pseudovalues) of the delayed objective. The pending objective values will be updated as long as the true values become available. This way, all candidate solutions will have the values of both the fast objective (true values) and the slow objective functions (pseudovalues), so that ranking and selection can be implemented at the frequency of evaluating the fast objective, denoted as $f_{f}$. Unsurprisingly, the performance of such a method heavily depends on the accuracy of the fitness approximation strategies. Subsequently, \cite{allmendinger2015multiobjective} gave a detailed description of MOPs with delayed objectives and suggested three schemes, termed \emph{Waiting}, \emph{Fast-first} and \emph{Interleaving schemes}. Actually \emph{Waiting} is the usual strategy that waits for all evaluations to be accomplished before selection is conducted, which therefore can be seen as the baseline method. By contrast, \emph{Fast-first} aims to make the best use of the fast evaluations. To achieve this, \emph{Fast-first} adopts a single-objective evolutionary algorithm (SOEA) to explore $f_{f}$ first and subsequently evaluates the slow objective of a set of solutions (the best and distinct solutions) obtained in the single-objective search. Therefore, the objective values available for $f_{s}$ are not fully exploited. Finally, \emph{Interleaving} scheme considers not only the evaluation budgets for different objectives, but also tries to co-ordinate the different evaluation times of the different objectives. Two implementations, \emph{Brood interleaving} and \emph{Speculative interleaving}, of {Interleaving} scheme were proposed in~\cite{allmendinger2015multiobjective}.
While \emph{Brood interleaving} focuses more on maintaining diversity, \emph{Speculative interleaving} uses an SOEA to optimize the fast objective when the slow objective is being evaluated.

The above studies exploit the fast objectives mainly with the help of evolutionary algorithms (EAs) and have not been extended to the context of SAEAs. However, as surrogates are a powerful tool to cope with limited budgets for function evaluations, it is a promising approach and deserves more attention, despite the fact that many factors need to be considered. \cite{chugh2018hkrvea} first attempted to extend a Kriging-assisted EA (K-RVEA) \citep{chugh2018surrogate} to bi-objective optimization where objective functions require heterogeneous evaluation times and proposed a variation of K-RVEA, called HK-RVEA containing an SOEA for selecting training data and K-RVEA for updating surrogates. Specifically, HK-RVEA employs an SOEA to optimize $f_{f}$ in the initialization, and genetic operators are adopted to generate additional solutions $\mathbf{X}^{a}$ for $f_{f}$ when waiting for the evaluation of new samples with $f_{s}$. It turns out that HK-RVEA performs well in cases where the ratio $\tau$ of the evaluation times for $f_{s}$ and $f_{f}$ is low. \cite{wang2020TSAEA} extended an SAEA by applying a parameter based transfer learning (TL) technique to bi-objective problems with heterogeneous objectives, called (T-SAEA). To transfer knowledge acquired during the search experience on $f_{f}$ for enhancing the performance of the Gaussian Process surrogate model for the slow objective, $GP_{s}$, T-SAEA identifies common decision variables related to both $f_{f}$ and $f_{s}$ by means of a filter-based feature selection. Subsequently, corresponding parameters in the surrogates can be shared to improve the quality of $GP_{s}$. \textcolor[rgb]{1,0,0}{In a follow-up work, \cite{wang2021transfer} proposed an instance-based TL for SAEA, called Tr-SAEA, where a hybrid domain adaptation method is introduced to align the source domain (the space of the fast objective $f_{f}$) and the target domain (the space of the slow objective $f_{s}$) in a Reproducing Kernel Hilbert Space, obtaining a mapping matrix. Then, additional data for the fast objective $y_{f}^{a}$ are mapped into the latent space to generate additional data for the slow objective $\mathbf{y}_{s}^{a}$, and the corresponding decision variables $\mathbf{X}_{s}^{a}$ of these data points are obtained by enforcing its objective values on $f_{s}$ to be close to $\mathbf{y}_{s}^{a}$ in the latent space. To leverage the unlabeled $\mathbf{X}_{s}^{a}$, a co-training method is introduced to boost the quality of the surrogate of $f_{s}$. However, the performance of Tr-SAEA highly depends on the dimension of the latent space to be specified, and an optimization algorithm is required to generate the transferable solutions in the decision space.}

\begin{figure}[ht]
\centering
\includegraphics[width=0.5\columnwidth]{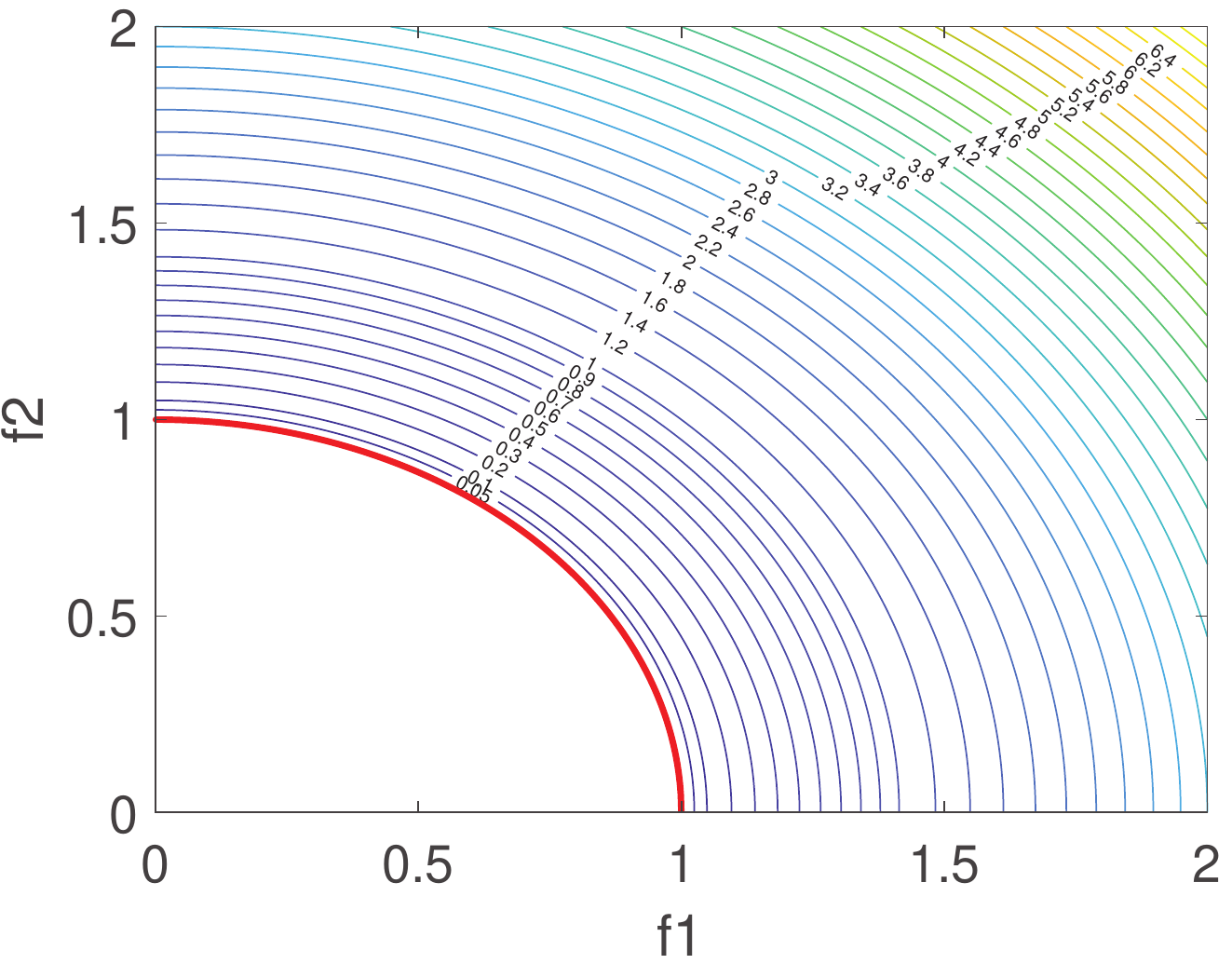}
\caption{An illustration of the functional relationship between objectives for DTLZ2, where the red line is the PF.}
\label{Fig.relationship}
\end{figure}

Given that sufficient data samples of search parameter and objective value pairs $(\mathbf{X}^{a},\mathbf{Y}_{f}^{a})$ are available for the computationally cheap fast objective $f_{f}$, it is natural to leverage the knowledge hidden in $(\mathbf{X}^{a},\mathbf{Y}_{f}^{a})$ to speed up the search process for $f_{s}$. This is warranted since  for solutions on the PF  there exists a functional relationship between $f_{f}$ and $f_{s}$ (usually a trade-off relation) \citep{deb1999multi,coello2007evolutionary}. In addition, such a functional relationship typically also approximately holds for solutions close to the PF. Take the bi-objective DTLZ2 test function \citep{deb1999multi} as an example, its Pareto front is defined by $f_1^2+f_2^2=1$. For any solution that is not on the Pareto front, an error function approximating the PF can be defined as $e (f_1, f_2) = f_1^2+f_2^2-1$. The contour plot of this error function is shown in Fig. \ref{Fig.relationship}, indicating that the closer a solution is to the PF, the smaller the error is, and thus the stronger the functional relationship between $f_1$ and $f_2$ will be.

In this paper, we aim to more efficiently solve bi-objective optimization problems with delayed objectives with the help of the additional instances obtained from the optimization of $f_{f}$ on the basis of SAEAs. To this end, we develop an instance TL scheme incorporated into a GP-assisted SAEA, which is a surrogate based interleaving method. The proposed algorithm, termed TC-SAEA, adopts a co-surrogate model to capture the underlying correlation between $f_{s}$ and $f_{f}$ on or near the PF supported by a transferable instance selection method to uncover useful knowledge. Here, GPs are employed as the surrogates and the adaptive acquisition function in \cite{wang2020adaptive} serves as the model management strategy. Similar to conventional SAEAs \citep{knowles2006parego,naujoks2005metamodel,zhang2009expensive}, a GP model is separately built for each objective function. The main contributions of the paper are as follows:
\begin{enumerate}
  \item A GP surrogate model is built for the fast and slow objective functions, respectively, to assist the search of a bi-objective optimization problem with non-uniform fitness evaluation latencies. To enhance the quality of the surrogate for the slow objective function, a co-surrogate model is constructed to learn the difference between the slow and fast objective values. The co-surrogate is employed to generate synthetic data ($\mathbf{X}^{a}, \mathbf{Y}_{s}^{'a}$) for training the surrogate of the slow objective, once the fast objective function of the additional solutions ($\mathbf{X}^{a}$) are evaluated.

  \item An instance-based TL mechanism is introduced to select transferable data from the synthetic data generated by the co-surrogate. In order to identify reliable samples from $(\mathbf{X}^{a}, \mathbf{Y}_{s}^{'a})$, a selection criterion based on the predicted value and the uncertainty of the prediction delivered by $GP_{s}$ is suggested.
\end{enumerate}

\section{Preliminaries}

In the following, we first introduce the related notations and assumptions of the problems under consideration.
The slow (or delayed/expensive) objective function is denoted as $f_{s}$, while the fast (or non-delayed/cheap) one is denoted as $f_{f}$, and the corresponding Gaussian process models (GPs) are denoted as $GP_{s}$ and $GP_{f}$, respectively. The ratio of the evaluation time of $f_{s}$ to that of $f_{f}$ is denoted by $\tau$. Here, we assume that $\tau$ is an integer larger than 1. Solutions that are evaluated by both $f_{f}$ and $f_{s}$ are denoted as $\mathbf{X}$. Consequently, the slow and fast objectives of $\mathbf{X}$ are denoted by $\mathbf{Y}_{s}$ and $\mathbf{Y}_{f}$, respectively. Due to the big difference in the evaluation time of the slow and fast objectives, to make full use of the time during which the slow objective is being evaluated, additional solutions, denoted by $\mathbf{X}^{a}$ can be evaluated for the fast objective function ($f_f$), which are denoted by $\mathbf{Y}_{f}^{a}$. Here, we assume that evaluation of the fast and slow objective functions can be done in parallel.

It is assumed that the computational time for building surrogates and for implementing the genetic operators is negligible compared to that for evaluating the expensive objectives. Consequently, the total computational time available for solving the problem under consideration is defined by the total budget of function evaluations. Specifically, the total budget is defined as the maximum number of function evaluations for the slow objective $f_{s}$, denoted as $FE_s^{max}$. Consequently, the maximum budget for the fast objective $FE_{f}^{max}$ equals $FE_{s}^{max}*\tau$.

\subsection{Kriging model}
Gaussian processes can be traced back to Kriging \citep{matheron1963principles} and have attracted increasing attention in machine learning. Meanwhile, GPs are also known as Kriging in efficient global optimization \citep{jones1998efficient} as well as in SAEAs \citep{jin2002gecco,allmendinger2017survey}. Therefore, these two terms will be used interchangeably in the rest of the paper. A major advantage of GPs models over other surrogate models is that they not only provide a prediction for the objective value but also a confidence level of the prediction, based on which we can obtain a prediction interval. The uncertainty information can be used in an acquisition function to select new samples for effectively updating the GPs.

A GP is a collection of random variables that have a joint multivariate Gaussian distribution for any finite set of inputs. Consider a training set that includes $N$ samples $(\mathbf{X},\mathbf{Y})$, where $\mathbf{X}=[\mathbf{x}^{1}, \mathbf{x}^{2}, \cdots,\mathbf{x}^{N}]^{T}$, $\mathbf{Y} =[y^{1}, y^{2}, \cdots,y^{N}]^{T}$, $i=1, 2,\cdots, N$. A GP can be specified by a mean and a covariance function, denoted by $\mu$ and $\sigma$. The covariance function is used to describe the correlation between $y^{i}$ and $y^{j}$ related to the distance between $\mathbf{x}^{i}$ and $\mathbf{x}^{j}$. In general, the squared exponential function with additional hyperparameters is employed to calculate the correlation:
\begin{equation}
d( \mathbf{x}^{i},\mathbf{x}^{j} ) =\sum_{k=1}^{m}\theta _{k} |x_{k}^{i}-x_{k}^{j} |^{p_{k}}
\end{equation}
\begin{equation}
{\rm Corr} (\mathbf{x}^{i}, \mathbf{x}^{j})={\rm exp} [ -d( \mathbf{x}^{i},\mathbf{x}^{j} ) ]
\end{equation}
where $p_{k}\in \left ( 0,1 \right )$ controls the smoothness of the function in terms of the $k$-th dimension, and $\theta _{k}> 0$ denotes the importance of this dimension. When there are $N$ training data, an $N\times N$ correlation matrix $\mathbf{C}$ will be obtained,
\begin{equation}
\mathbf{C=}
\begin{pmatrix}
{\rm Corr} ( \mathbf{x}^{1},\mathbf{x}^{1}  ) & \cdots &{\rm Corr} ( \mathbf{x}^{1},\mathbf{x}^{N} )\\
\vdots & \ddots & \vdots\\
{\rm Corr} ( \mathbf{x}^{N},\mathbf{x}^{1})& \cdots &{\rm Corr}( \mathbf{x}^{N},\mathbf{x}^{N} )
\end{pmatrix}
\end{equation}
As a result, the hyperparameters will determine a GP model, which can be estimated by maximizing the following likelihood function,
\begin{equation}
\psi \left ( \theta _{1},\cdots, \theta _{N},p_{1},\dots,p_{N} \right )=-\frac{1}{2}\left ( N\ln \sigma ^{2}+\ln det\left ( \mathbf{C} \right ) \right )
\end{equation}

Therefore, the estimates $\hat{\mu}$ and $\hat{\sigma}^{2}$ for the true values $\mu$ and $\sigma$ will be obtained
\begin{equation}
\hat{\mu }=\frac{\mathbf{1}^{T}\mathbf{C}^{-1}\mathbf{y}}{\mathbf{1}^{T}\mathbf{C}^{-1}\mathbf{1}}
\end{equation}
\begin{equation}
\hat{\sigma  }^{2}=\frac{(\mathbf{y}-\mathbf{1}\hat{\mu } )^{T}\mathbf{C}^{-1} (\mathbf{y}-\mathbf{1}\hat{\mu } )}{N}
\end{equation}
where $\mathbf{1}$ denotes an $N\times1$ column vector of ones.
Based on the given parameters, the GP model can predict the mean value together with a variance for a new data point $\mathbf{x}^{new}$ (which corresponds to the decision variables of a candidate solution when the GP is used as a surrogate),
\begin{equation}
f ( \mathbf{x}^{new} )=\hat{\mu }+\mathbf{r}^{T}\mathbf{C^{-1}} (\mathbf{y}-\boldsymbol{1}\hat{\mu }  )
\end{equation}
\begin{equation}
\hat{\sigma}(\mathbf{x}^{new})^{2}=\hat{\sigma }^{2}[1-\mathbf{r}^{T}\mathbf{C}^{-1}\mathbf{r}+\frac{(1-\mathbf{r}^{T}\mathbf{C}^{-1}\mathbf{r})^{2}}{\mathbf{1}^{T}\mathbf{C}^{-1}\mathbf{1}}]
\end{equation}
where $\mathbf{r}= ( {\rm Corr}(\mathbf{x}^{new},\mathbf{x}^{1}),\cdots, {\rm Corr}(\mathbf{x}^{new},\mathbf{x}^{N}) )^{T}$ presents a correlation vector between $\mathbf{x}^{new}$ and each element $\mathbf{x}^{i}$ in $\mathbf{X}$.

\subsection{Acquisition Function}
In SAEAs, it is of paramount importance to select the right candidate solutions to be evaluated using the real objective function so that the quality of the surrogates can be improved as much as possible and that the surrogates are able to efficiently guide the evolutionary search to the optimum. This is known as model management, which plays a key role in SAEAs \citep{jin2000evolutionary, allmendinger2017survey}.

In GP-assisted optimization such as Bayesian optimization, model management is done by optimizing an acquisition function (AF). AFs can generally be divided into three categories, improvement-based, information-based and optimistic\citep{noe2018new}. Improvement-based methods aim to select solutions (samples) with the highest probability of improving upon the best observed sample so far, such as probability of improvement (PI) \citep{kushner1964new} and expected improvement (EI) \citep{jones1998efficient}. Instead of querying at solutions where we expect to obtain promising fitness values, information-based methods focus on finding out new solutions containing more information about the location of the optimum. For example, entropy search (ES) \citep{hennig2012entropy} and predictive entropy search (PES) \citep{hernandez2014predictive} both query at solutions with the largest reduction of uncertainty of the surrogate. Optimistic methods design acquisition functions based on an optimistic attitude towards uncertainty. A representative one is the lower confidence bound (LCB) \citep{cox1992statistical} combining the uncertainty with the predicted objective values \citep{liu2012comparison}. An adaptive acquisition function based on LCB is introduced by \cite{wang2020adaptive} to tune the trade-off parameter according to the search dynamics, achieving a better balance between the convergence and diversity. In this work, we adopt this acquisition function to select new samples to be evaluated by the original objective functions.

\subsection{Instance-based Transfer Learning}
Among various TL techniques, instance-based transfer learning is intuitively appealing: although the source domain data cannot be reused directly, there are certain parts of the data that can still be reused together with a few labeled data in the target domain \citep{pan2009survey}. In the machine learning community, many instance-based transfer learning approaches have been proposed, most of which adopts instance weighting strategies. For example, TrAdaBoost \citep{dai2007boosting} combines the labeled source-domain and target-domain instances together as the training data, and adjusts the weights of instances to reduce the negative effects of the source domain. In CP-MDA \cite{chattopadhyay2012multisource}, source classifiers label the unlabeled target data using a weighting scheme based on the similarities between the conditional probabilities of the source and target domain data.

Measuring the relatedness or similarity between the source and target domains is a vital issue for studying transferability \citep{duan2012learning}. For example, in the context of machine learning, \cite{yang2015learning} employed a directed cyclic network to denote transferred weights from a source domain to a target domain, and then the relatedness between the given domains is evaluated through the transfer weights. \cite{gong2012geodesic} introduced a metric called rank of domain (ROD) to rank a list of source domains by evaluating each source domain in terms of the degree of overlap and similarity to the target domain. Similarly, in the context of evolutionary computation, the similarity between tasks is measured based on the KL-divergence of the distributions of the training data sets in \citep{huang2019surrogate}. \cite{min2017multiproblem} proposed a model-based transfer stacking approach to combining multiple surrogates together with a linear meta-regression model, where the meta-regression coefficients indicate the similarities between the source and target optimization tasks.

\section{Proposed Algorithm}
We give the related notations in the following, before introducing the proposed algorithm.

As we mentioned in Section I, surrogate models $GP_{s}$ and $GP_{f}$ are constructed for $f_{s}$ and $f_{f}$, respectively. $D_{s}={(\mathbf{X}, \mathbf{Y}_{s})}$ is defined as the training data set for $GP_{s}$, while $D_{f}=\left\{(\mathbf{X}, \mathbf{Y}_{f}), (\mathbf{X}^{a}, \mathbf{Y}_{f}^{a})\right\}$ is defined as the training data set for $GP_{f}$. Apart from $GP_{s}$ and $GP_{f}$, we construct a co-surrogate model $GP_{c}$ to approximate the difference (denoted as function $f_{c}$) between $f_{s}$ and $f_{f}$, indirectly describing the relationship between $f_{s}$ and $f_{f}$ on or near the PF. Specifically, the input-output pairs ${(\mathbf{X}, \mathbf{Y}_{c})}$ are defined as the training data set $D_{c}$ for $GP_{c}$, where $\mathbf{Y}_{c}=\mathbf{Y}_{s}-\mathbf{Y}_{f}$. Note that the training data for $GP_{c}$ are promising solutions, as they are selected according to the AF. This way, for the additional solutions $\mathbf{X}^{a}$ (defined in Section I), the synthetic objective values $\mathbf{Y}_{s}^{'a}$ of $f_{s}$ can be calculated by $\mathbf{Y}_{c}^{a}+\mathbf{Y}_{f}^{a}$, where $\mathbf{Y}_{c}^{a}$ is predicted by $GP_{c}$. The input-output pairs ${(\mathbf{X}^{a}, \mathbf{Y}_{s}^{'a})}$ are defined as an auxiliary data set $D_{a}$ for $f_{s}$. To select the most reliable data in the auxiliary data set $D_{a}$, a selection criterion is designed by calculating a confidence interval $CI$. Specifically, for the additional solutions $\mathbf{X}^{a}$, $GP_{s}$ predicts objective values $\mathbf{Y}_{s}^{a}$ together with their variances $\boldsymbol{\sigma}_{s}^{a}$. Subsequently, $CI=\mathbf{Y}_{s}^{a} \pm \boldsymbol{\sigma}_{s}^{a}$ is defined as a selection threshold. Then the selected transferable data set is defined as $D_{t}={(\mathbf{X}^{t}, \mathbf{Y}_{s}^{t})}$.

\begin{figure}[ht]
\centering
\includegraphics[width=0.8\columnwidth]{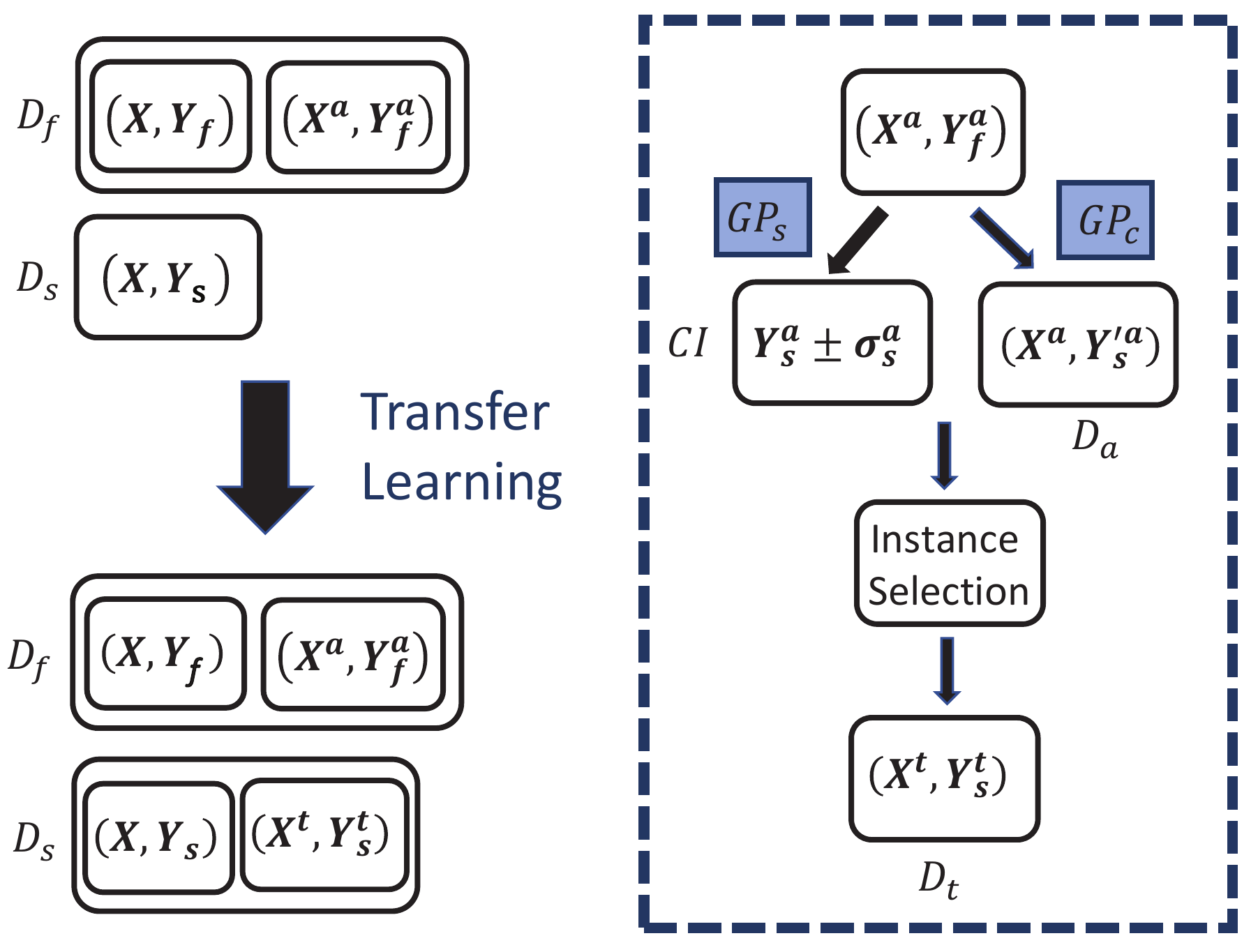}
\caption{An illustration of the proposed transfer learning scheme. Left: Augmentation of $D_s$ using the proposed transfer learning scheme. Right: A diagram for the transfer learning method adopted in this work.}
\label{Fig.1}
\end{figure}

For solving MOPs with delayed objectives, the algorithm should be able to quickly find a set of diverse and well converged solutions due to the limited budget of function evaluations. While the previous studies only focus on the utilization of per-objective evaluation budgets and straightforward instance transmission, the main motivation of this work is to augment the data set $D_{s}$ by transferring knowledge acquired from the additional solutions so as to accelerate the convergence towards the Pareto front. To achieve this, there are two key challenges. One is how to generate synthetic objective values $\mathbf{Y}_{s}^{'a}$ to the additional solutions $\mathbf{X}^{a}$ when the observed data for $f_{s}$ is scarce; the other is how to identify the most reliable one from the auxiliary data set ${(\mathbf{X}^{a}, \mathbf{Y}_{s}^{'a})}$ as the transferable data set $D_{t}$. To address these issues, we propose here a transfer learning scheme to be incorporated into a GP-based SAEA, illustrated in Fig. \ref{Fig.1}. As shown in the left figure, an augmented data set $D_{s}$ can be obtained by transferring some useful data from the auxiliary data set ${(\mathbf{X}^{a}, \mathbf{Y}_{s}^{'a})}$. The right figure depicts the workflow diagram for the proposed transfer learning scheme. Specifically, a co-surrogate $GP_{c}$ is built to approximate the functional relationship between the $\mathbf{X}$ and $(\mathbf{Y}_{s}-\mathbf{Y}_{f})$ to bridge the gap between $f_{s}$ and $f_{f}$. This way, synthetic values $\mathbf{Y}_{s}^{'a}$ for the additional solutions can be calculated by adding $\mathbf{Y}_{c}^{a}$ to $\mathbf{Y}_{f}^{a}$, thereby obtaining the auxiliary data set $D_{a}={(\mathbf{X}^{a}, \mathbf{Y}_{s}^{'a})}$. However, some synthetic values may be detrimental if the predicted values are extremely inaccurate. Hence, we can take advantage of the confidence level $CI=\mathbf{Y}_{s}^{a} \pm \boldsymbol{\sigma}_{s}^{a}$ given by $GP_{s}$ as a selection threshold to filter out the unreliable auxiliary samples.

\subsection{Algorithm Framework}
\begin{figure*}[ht]
\centering
\includegraphics[width=1\textwidth]{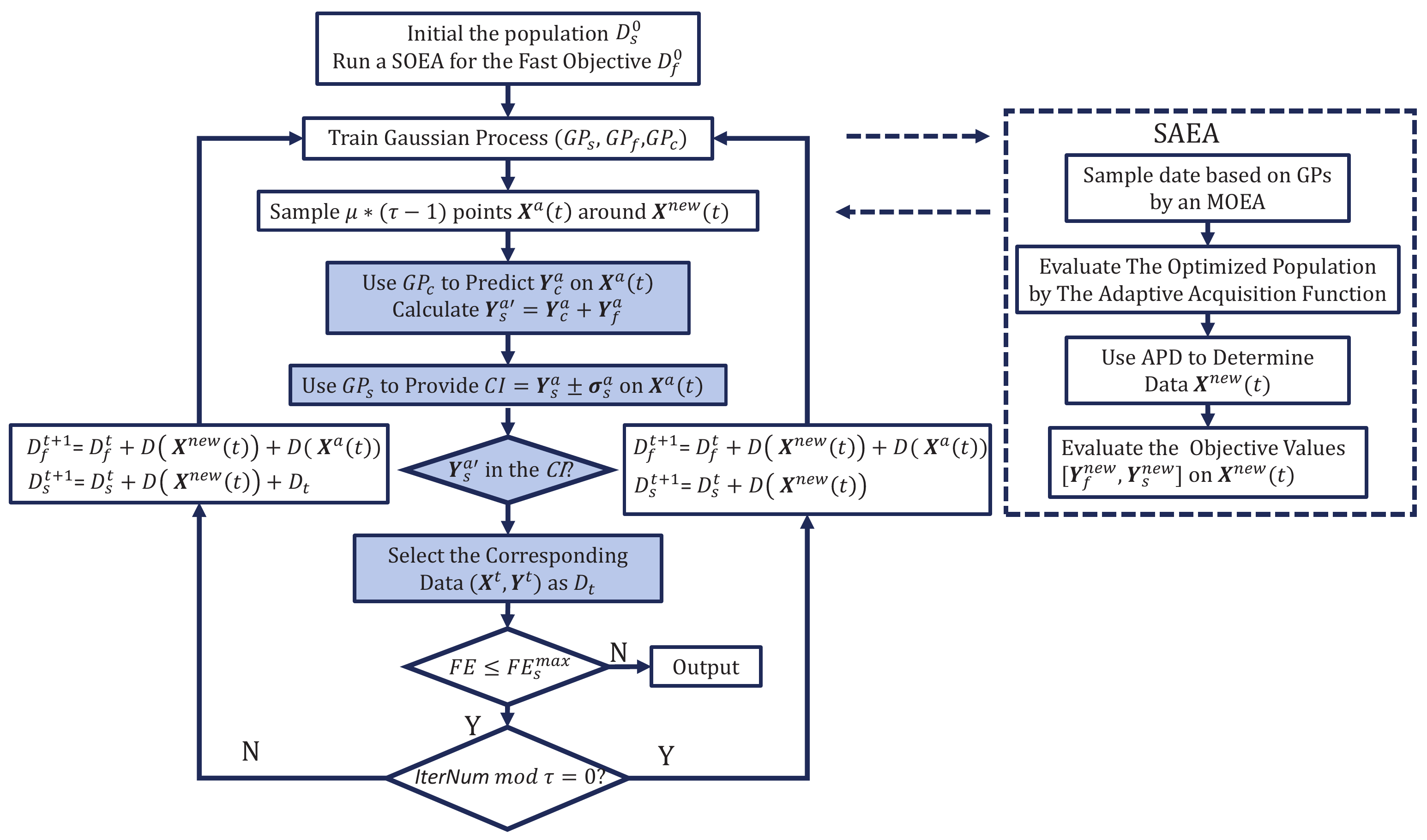}
\caption{The framework of TC-SAEA.}
\label{Fig.2}
\end{figure*}

The framework of the proposed TC-SAEA \footnote{The source code is available at https://github.com/xw00616/TC-SAEA.git.} is shown in Fig. \ref{Fig.2} and its pseudo code is outlined in Algorithm \ref{Algorithm 1}. In the following, we detail the main components of TC-SAEA.

\begin{algorithm}[htb]\footnotesize{
\renewcommand{\algorithmicrequire}{\textbf{Input:}}
\renewcommand{\algorithmicensure}{\textbf{Output:}}
\caption{The framework of TC-SAEA} \algblock{Begin}{End}
\label{Algorithm 1}
\begin{algorithmic}[1]
\Require $FE_{s}^{max}$: the maximum number of the slow objective function evaluations; $\tau$: the ratio of the evaluation times between the slow and fast objectives; $u$: the number of new samples for updating the GP; $w_{max}$: the maximum number of generations before updating GPs;\textcolor[rgb]{1,0,0}{ $N_{max}$: the maximum number of training data points.}
\Ensure The optimized solutions in $D_{s}$;
\State Initialization: Use the Latin Hybercube Sampling method to sample an initial population $\mathbf{X}$; calculate $\mathbf{Y}_{s}$, $\mathbf{Y}_{f}$ and the difference $\mathbf{Y}_{c}=\mathbf{Y}_{s}-\mathbf{Y}_{f}$. Set $D_{s}^{0}={(\mathbf{X}, \mathbf{Y}_{s})}$, $D_{f}^{0}={(\mathbf{X}, \mathbf{Y}_{f})}$ and $D_{c}^{0}={(\mathbf{X}, \mathbf{Y}_{c})}$. Run an SOEA to optimize $f_{f}$ and save data in $D_{f}^{0}$; set $FE_{s}=size(D_{s}^{0})$ and $IterNum=1$.\;
 \While{$FE_{s}\leqslant FE_{s}^{max}$}
  \State \textcolor[rgb]{1,0,0}{Limit the size of the training data to $N_{max}$ and} train $GP_{f}$ with the training data $D_{f}$ for $f_{f}$; \;
  \State \textcolor[rgb]{1,0,0}{Limit the size of the training data to $N_{max}$ and} train $GP_{s}$ with the training data $D_{s}$ for $f_{s}$;\;
  \State \textcolor[rgb]{1,0,0}{Limit the size of the training data to $N_{max}$ and} train $GP_{c}$ with the training data $D_{c}$ for $f_{c}$;\;
     \While{$w\leqslant w_{max}$}
       \State //Using the surrogate in the EA//
       \State  Run an MOEA to find samples for updating GPs;\;
       \State $w=w+1$;\;
      \EndWhile
  \State Use the adaptive acquisition function to evaluate the individuals in the optimized population;\;
  \State Use the angle-penalized distance (APD) to determine $u$ new solutions $\mathbf{X}^{new}$ to be evaluated with both $f_{s}$ and $f_{f}$, and add new data to $D_{s}$;\;
  \State Sample $u*(\tau-1)$ additional solutions $\mathbf{X}^{a}$ around $\mathbf{X}^{new}$ to be evaluated by $f_{f}$, and add new data $\mathbf{X}^{new}$ and $\mathbf{X}^{a}$ to $D_{f}$;\;
  \State //Using the co-surrogate to select the transferable instances//
  \State Use $GP_{c}$ to predict the values of $\mathbf{Y}_{c}^{a}$ on $\mathbf{X}^{a}$ and calculate the synthetic values $\mathbf{Y}_{s}^{'a}=\mathbf{Y}_{c}^{a}+\mathbf{Y}_{f}^{a}$;\;
  \State Use $GP_{s}$ to predict the objective values $\mathbf{Y}_{s}^{a}$ and the uncertainty $\mathbf{\sigma}_{s}^{a}$ on $\mathbf{X}^{a}$, calculate the confidence interval $CI=\mathbf{Y}_{s}^{a}\pm \boldsymbol{\sigma}_{s}^{a}$ ;\;
  \If {$\mathbf{Y}_{s}^{'a} \in CI$}
  \State The corresponding data ${(\mathbf{X^{t}}, \mathbf{Y}_{s}^{t})}$ is transferable and saved in $D_{t}$;\;
  \EndIf
  \If {$IterNum mod \tau==0$}
  \State Train $GP_{s}$ with $D_{s}$
  \Else
  \State Train $GP_{s}$ with $D_{s}+D_{t}$
  \EndIf
  \State Update $FE_{s}=FE_{s}+u$, $IterNum=IterNum+1$;
\EndWhile
 \State  Return the optimized solutions in $D_{s}$;
\end{algorithmic}}
\end{algorithm}

The algorithm starts with using the Latin Hybercube Sampling (LHS) method to generate an initial population $\mathbf{X}$, and the corresponding objective values $\mathbf{Y}_{s}$ and $\mathbf{Y}_{f}$ on $\mathbf{X}$ are calculated with the true objective functions $f_{s}$ and $f_{f}$. The difference between the two objective values $\mathbf{Y}_{c}=\mathbf{Y}_{s}-\mathbf{Y}_{f}$ can be obtained for $\mathbf{X}$. Let $D_{s}={(\mathbf{X}, \mathbf{Y}_{s})}$, $D_{f}=(\mathbf{X}, \mathbf{Y}_{s})$ and $D_{c}={(\mathbf{X}, \mathbf{Y}_{c})}$. In the initialization, additional function evaluations for $f_{f}$ are available when waiting for the time-consuming evaluation of $f_{s}$. There are different possible ways of exploiting the additional evaluations of the fast function, and in this work we follow the \emph{Interleaving scheme} in \citep{allmendinger2015multiobjective}, where an SOEA is employed to optimize the fast objective function $f_{f}$ and the obtained data is saved in $D_{f}$. Subsequently, $GP_{s}$, $GP_{f}$ and $GP_{c}$ are trained with data sets $D_{s}$, $D_{f}$ and $D_{c}$, respectively. \textcolor[rgb]{1,0,0}{To reduce the computational complexity, we used the same method in ParEGO \citep{knowles2006parego} to limit the size of the training data. Specifically, when the number of training data exceeds a predefined upper bound $N_{max}$, a selection strategy will be triggered: the first half $N_{max}/2$ training data are the best ones sorted according to the objective values, and the rest half are randomly sampled from the training dataset without replacement.}

Here, we use RVEA \citep{cheng2016reference} as the baseline MOEA to perform multi-objective optimization assisted by the surrogate models $GP_{s}$ and $GP_{f}$ for function evaluations, instead of directly searching on the real objective functions $f_{s}$ and $f_{f}$. In every 20 generations, the adaptive acquisition function (AFF) \citep{wang2020adaptive} is then applied to evaluate each individual in the population, subsequently, the angle-penalized distance (APD) selection criterion in RVEA is adopted to identify a predefined number of new samples $\mathbf{X}^{new}$ according to the AFF. Note that samples in $\mathbf{X}^{new}$ are evaluated with the real objective functions $f_{s}$ and $f_{f}$, thereby involving the time-consuming function evaluation $f_{s}$ again. Hence, different from conventional SAEAs, the fast objective function can be called more times to evaluate $u*(\tau-1)$ (where $\tau$ is an integer larger than 1 as defined in Section I) additional new solutions $\mathbf{X}^{a}$ sampled using LHS around $\mathbf{X}^{new}$.

It is apparent that there is a considerable amount of observed data for $f_{f}$, while the data for $f_{s}$ is inadequate due to the different evaluation times. As a result, the convergence of $f_{s}$ will be slow and a bias towards the fast objective function will be introduced in the search \citep{allmendinger2015multiobjective}, posing challenges for SAEAs to efficiently solve such problems. Thus, it would make sense to augment the data for $f_s$ to alleviate the problem in training $GP_{s}$ caused by a limited evaluation budget for $f_{s}$. Therefore, a transfer learning scheme is proposed to make use of the knowledge extracted from the additional solutions $\mathbf{X}^{a}$. The proposed scheme includes two main components: the GP co-surrogate model, denoted as $GP_{c}$, to indirectly describe the functional relationship between the two objective functions, and the transferable instance selection criterion to determine transferable data from the auxiliary data set $D_{a}$. Specifically, given the additional solutions $\mathbf{X}^{a}$, an auxiliary data set $D_{a}={(\mathbf{X}^{a}, \mathbf{Y}_{s}^{'a})}$ can be attained, where the synthetic objective values are calculated using the co-surrogate and the true objective values $\mathbf{Y}_{f}$ of $f_{f}$. Note that it is non-trivial to build effective surrogates, which means that the synthetic data generated by $GP_{c}$ may be subject to large errors. Consequently, the instance selection method is executed to identify transferable data $D_{t}$ from the auxiliary data set to promote the performance of TC-SAEA in terms of both optimization quality and convergence speed. The selection criterion is designed by calculating $CI=\mathbf{Y}_{s}^{a} \pm \boldsymbol{\sigma}_{s}^{a}$, where $\mathbf{Y}_{s}^{a}$ and $\boldsymbol{\sigma}_{s}^{a}$ are provided by $GP_{s}$. If the synthetic objective values $\mathbf{Y}_{s}^{'a}$ are within the bound $CI$, the corresponding auxiliary data is selected as transferable data $D_{t}$.

For $\tau$ iterations, $GP_{s}$ is updated once using the data in $D_{s}$ without $D_{t}$; otherwise, $GP_{s}$ is trained using both the transferable data set $D_{t}$ and the observed data $D_{s}$ evaluated with the real objective function $f_{s}$. Augmenting the data for $f_{s}$ with the help of the transfer learning scheme can alleviate the problem resulting from the insufficient data for $f_{s}$.

\subsection{Co-surrogate Model}
\label{sec:co-surr}
As shown in instance-based transfer learning, it is a promising way to utilize the source-domain instances, i.e. $(\mathbf{X}^{a},\mathbf{Y}_{f}^{a})$ to augment the training data for $GP_s$. As we introduced before, the key question is how to estimate the functional relationship between the two objectives. In bi-objective optimization in which the objectives have different evaluation times, there are only two objective functions and many solutions evaluated with each objective function may be the same due to the population-based SAEA. As a result, the slow objective function $f_{s}$ only has one auxiliary task $f_{f}$ and the transferable knowledge is limited to the additional data achieved on the fast objective $f_{f}$; besides, it is impracticable to use the objective values $\mathbf{Y}_{s}$ and $\mathbf{Y}_{f}$ (column vectors) to directly model the relationship between $f_{s}$ and $f_{f}$. Therefore, the previously discussed TL techniques are not well suited for the problem considered in this work.

To address this issue, we build a regression model, called co-surrogate model $GP_{c}$, to estimate the underlying relationship between $f_{s}$ and $f_{f}$. Note that there is a certain functional relationship between the two objectives, which may only loosely hold in most of the search space but does hold for all Pareto optimal solutions \citep{deb1999multi,coello2007evolutionary}. In other words, the closer to the Pareto front, the stronger the functional relationship between $f_f$ and $f_s$ will be. Recall that the data for training the co-surrogate $GP_{c}$ are promising solutions, as they are all selected according to the acquisition function. Therefore, $GP_{c}$ can describe the underlying relationship between the two objectives on/close to the PF by learning from ${(\mathbf{X}, \mathbf{Y}_{c})}$, where $\mathbf{Y}_{c}$ is the difference between the slow and fast objectives as defined in Equation (\ref{eq:diff}). \textcolor[rgb]{1,0,0}{It should be stressed that $GP_{c}$ is an approximation of $\mathbf {Y_{c}}$, which is a function of $\mathbf{X}$ that can be linear or nonlinear. More discussions about the co-surrogate will be provided in Section \ref{sec:co-surrogate}.}

Given $N$ inputs $\mathbf{X}=[\mathbf{x}^{1}, \mathbf{x}^{2},\cdots,\mathbf{x}^{N}]^{T}$ evaluated on both the fast and slow objectives $[\mathbf{Y_{f}}, \mathbf{Y_{s}}]$, where $\mathbf {Y_{f}}=[y_{f}^{1}, y_{f}^{2},\cdots,y_{f}^{N}]^{T}$ and $\mathbf {Y_{s}}=[y_{s}^{1}, y_{s}^{2}, \cdots,y_{s}^{N}]^{T}$, the difference of the two objectives can be calculated as follows:
\begin{equation}
\label{eq:diff}
\mathbf {Y_{c}}=[y_{s}^{1}-y_{f}^{1}, y_{s}^{s}-y_{f}^{2},\cdots,y_{s}^{N}-y_{f}^{N}]^{T}
\end{equation}
We can build a co-surrogate by training a GP with the input-output pairs $(\mathbf{X},\mathbf {Y_{c}})$. Compared with $GP_s$, $GP_c$ learns from the solutions evaluated on both objectives from a different perspective since the difference between $f_s$ and $f_f$ is the output. Once the additional solutions $\mathbf{X}^{a}$ associated with $\mathbf{Y}_{f}^{a}$ are obtained, the co-surrogate $GP_{c}$ can predict the difference $\mathbf{Y}_{c}^{a}$ between $f_{s}$ and $f_{f}$ on $\mathbf{X}^{a}$, then the synthetic values of $f_{s}$ on $\mathbf{X}^{a}$ are determined as follows:
\begin{equation}
\mathbf{Y}_{s}^{'a}=\mathbf{Y}_{c}^{a}+\mathbf{Y}_{f}^{a}
\end{equation}
Therefore, the auxiliary data set for $f_{s}$ is defined as $D_{a}={(\mathbf{X}^{a},\mathbf{Y}_{s}^{'a})}$. A major advantage of using $GP_{c}$ is that the synthetic data generated for $GP_{s}$ can make use of the true objective values of $f_{f}$ and may improve the quality of the synthetic values of $f_{s}$, which is confirmed in Section \ref{Ablation Studies}.

\subsection{Transferable Instance Selection \label{selection}}
The co-surrogate model is constructed to connect the two objective functions by modelling the underlying relatedness so that the auxiliary data set can be generated and used to augment the training dataset for $GP_{s}$. Similar to the intuitive approaches in \citep{allmendinger2013hang, chattopadhyay2012multisource}, the synthetic values or pseudo labels can be utilized directly to augment the training dataset. Typically, the performance of these proposed algorithms may be improved by directly leveraging such synthetic objective values or pseudo labels, which, however, heavily depends on the quality of the synthetic objective values and pseudo labels. It is well recognized that identifying transferable data plays a vital role in transfer learning to alleviate negative transfer \citep{pan2009survey}. In the auxiliary data set $D_{a}={(\mathbf{X}^{a}, \mathbf{Y}_{s}^{'a})}$ obtained so far, the accuracy of the synthetic objective values $\mathbf{Y}_{s}^{'a}$ cannot be guaranteed, because these data are generated by only considering the underlying relationship between $f_{s}$ and $f_{f}$.
 For example, if the instances are incorrectly predicted, the corresponding synthetic values or pseudo labels are likely to conflict with the real values/labels, thereby misleading the training of the model and undermining the model's performance.
Hence, it is expected to transfer the reliable instances to ensure positive transfer learning and filter out the inaccurate synthetic instances. Besides, considering the fact that the co-surrogate is not as accurate as we expect and the functional relationship of the two objectives may be weak, it is necessary to carefully identify which instances can be transferred to train the surrogate model $GP_{s}$ and further improve the performance of TC-SAEA in terms of convergence and diversity.

\begin{figure}[ht]
\centering
\includegraphics[width=0.8\columnwidth]{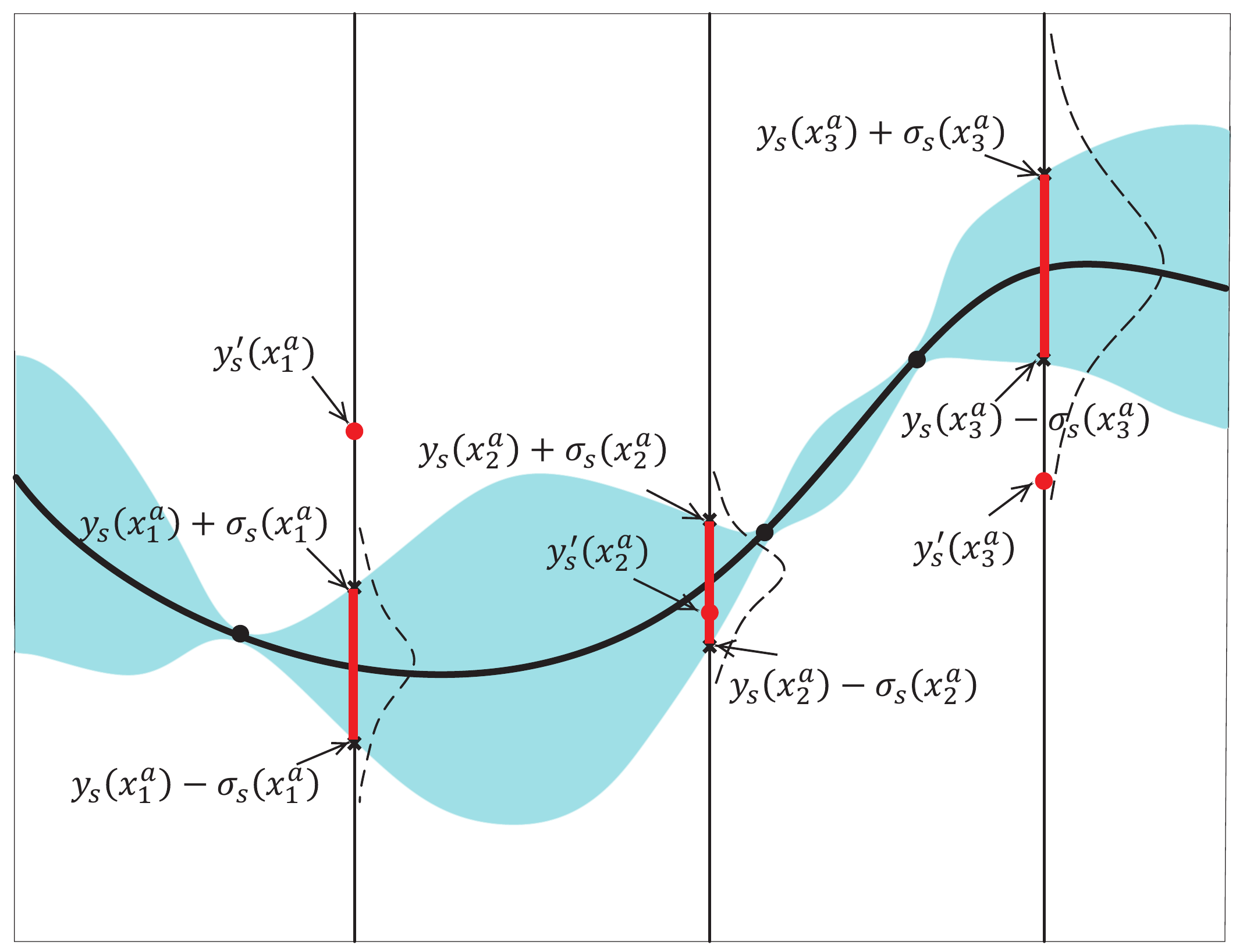}
\caption{An illustrative example of transferable instance selection for a $1D$ Gaussian process $GP_{s}$ with three training data points. For the three new data points ($x^a_1$, $x^a_2$ and $x^a_3$), the corresponding confidence intervals (as shown in thick red line) and the synthetic values (red points) are obtained by $GP_{s}$ and $GP_{c}$, respectively. The solid line and the shaded area indicate the mean and confidence intervals estimated with the GP model $GP_{s}$. Here, $y_s^{'}(x_2^a)$ is considered to be reliable.
}
\label{Fig.3}
\end{figure}

To select a subset of the data in $D_a$ as useful information for training $GP_s$, the confidence level predicted by $GP_{s}$ is adopted to assess the transferability of the auxiliary data $D_{a}={(\mathbf{X}^{a}, \mathbf{Y}_{s}^{'a})}$. Note that the GP model $GP_{s}$ for $f_{s}$ is initialized by training with $D_{s}={(\mathbf{X}, \mathbf{Y}_{s})}$, where $\mathbf{Y}_{s}$ is the real values of $f_{s}$. Therefore, it is not unlikely that the predicted objective value of an instance in $D_{a}$ is inconsistent with the prediction of $GP_{s}$. The hypothesis we make here is that the prediction given by the co-surrogate $GP_c$ is unreliable if the predicted value is out of the boundary predicted by $GP_s$. Specifically, for the additional solutions $\mathbf{X}^{a}$, $GP_{s}$ provides its prediction in terms of the mean objective value $\mathbf{Y}_{s}^{a}$ and the standard deviation $\boldsymbol{\sigma}_{s}^{a}$. Accordingly, the confidence interval is defined as
\begin{equation}
CI=\mathbf{Y}_{s}^{a} \pm \boldsymbol{\sigma}_{s}^{a}
\end{equation}
Therefore, if a synthetic value in $\mathbf{Y}_{s}^{'a}$ is within the range of $CI$, the associated instance is believed reliable and will be saved in the transferable data set $D_{t}$. On the contrary, if not, the synthetic value is regarded as unreliable in terms of the current GP model $GP_{s}$ for $f_{s}$. Fig. \ref{Fig.3} presents an illustrative example of how the transferable instance selection works. A $1D$ GP model $GP_{s}$ is trained with three training data, the GP model can then predict the mean fitness and the uncertainty of any new data point. For instance, for a given additional solution $x_1^{a}$, the synthetic value $y_{s}^{'}(x_1^{a})$ can be calculated with the help of $GP_{c}$, the mean objective value $y_{s}(x_1^{a})$, and the corresponding variances $\sigma_{s}^{a}(x_1^{a})$ can be predicted by $GP_{s}$. Since the predicted objective value $y_{s}^{'}(x_1^{a})$ is out of the interval $CI$ (indicated by the thick red line), this synthetic data will be considered as unreliable and will not be added to $D_{t}$. It is worth mentioning that, from the exploration perspective, selecting instances whose synthetic values are out of $CI$ may facilitate more exploratory search. However, note that the instances with the synthetic values are adopted to train the surrogate $GP_{s}$ rather than being evaluated with the true objective function $f_{s}$ before updating $GP_{s}$. Therefore, the key issue is to guarantee the reliability of the synthetic values to reduce the chance of misleading the search assisted by the surrogate.
\section{Comparative Studies}
In this section, two groups of bi-objective test problems and a test function with a controllable correlation between objectives are employed to examine the efficiency of the proposed algorithm and we assume $f_2$ (objective 2) in the benchmark problems is the slow objective $f_{s}$, as this work focuses on bi-objective optimization problems with different evaluation times. We run each algorithm under comparison on each benchmark problem for 20 independent times, and the inverted generational distance \citep{bosman2003balance} and hypervolume \citep{while2006faster} are used as the performance measures. The Wilcoxon rank sum test is adopted to compare the mean results achieved by TC-SAEA and other algorithms under comparison at a significance level of 0.05. Symbol "(+)" and "(--)" indicate that the proposed algorithm performs significantly better or significantly worse than the compared algorithm, respectively, while "($\approx)$" means there is no significant difference between them.

In the following section, a brief introduction to the test problems and the performance indicators is given at first, followed by a description of the details of the experimental settings concerning the optimization algorithms. The comparative experimental results of TC-SAEA and some state-of-the-art delay-handling methods are presented and discussed. To get a deeper insight into the proposed strategies, TC-SAEA is also compared with its three variants. Then, the impact of the correlation between objectives on the performance is investigated. Lastly, a sensitivity analysis is given.

\subsection{Test Problems}
Numerical experiments are conducted on sixteen bi-objective benchmark problems, nine of them taken from the DTLZ test suite \citep{deb2002scalable}, including DTLZ1 to DTLZ7 and two modified counterparts (DTLZ1a and DTLZ3a) of DTLZ1 and DTLZ3, and the remaining seven (UF1-UF7) from the UF test suite \citep{zhang2008multiobjective}. Specifically, DTLZ1a and DTLZ3a are designed to reduce the complexity to a reasonable level by changing the multi-model $g$ function used in DTLZ1 and DTLZ3. The $g$ function, given in the following, is suggested to control the ruggedness of DTLZ1 and DTLZ3 \citep{deb2002scalable},
\begin{equation}
g=100[5+\sum_{i\in {1,\dots,n}}(x_{i}-0.5)^{2}-{\rm cos}(20\pi(x_{i}-0.5))], i=1,\cdots,n.
\end{equation}
where $n$ denotes the number of decision variables. As indicated in~\citep{yang2019off}, $20\pi$ within the cosine term triggers excessively ruggedness. Consequently, $2\pi$ is adopted to reduce the complexity to a reasonable level. As recommended in \citep{deb2002scalable}, the number of decision variables for the DTLZ test instances is set to $n=M+K-1$, where $K=5$ is adopted for DTLZ1 and DTLZ1a, $K=10$ is used for DTLZ2 to DTLZ6 as well as DTLA3a, and $K=20$ is employed in DTLZ7. $M$ represents the number of objectives, here $M=2$. The number of decision variables for UF test suite is set as 30.

To demonstrate the sensitivity of the proposed algorithm on the correlation among objectives, a continuous version of the mapped OneMax problem (cm-OneMax) in \citep{chugh2018surrogate} is used. Let $n(\mathbf{x})$ and $n(\mathbf{x}^{map})$ be the sum of all variable values in a decision vector $\mathbf{x}$ and $\mathbf{x}^{map}$, respectively. Here, $\mathbf{x}^{map}$ is a mapped version of $\mathbf{x}$. The cm-OneMax problem is defined as:
\begin{equation}
f=\left(f_{1}, f_{2}\right)=\left(n(\mathbf{x}), n(\mathbf{x}^{map})\right)
\end{equation}
where $x^{map}_{i}=\left|x_{i}-map_{i}\right|, i=1,\cdot, n$, $n$ is the number of decision variables and is set to 10; $map_i \in [0,1]$ is set independently for each decision variable by flipping a coin biased by the degree of correlation $corr \in [-1,1]$ desired. For example, $corr=0$ means no correlation between the objectives, while $corr=1$ and $corr=-1$ mean maximal positive and maximal negative correlation, respectively. Given the degree of correlation, $map_i$ is set to zero with a probability of $(1+corre)/2$.

\subsection{Performance Metrics}
The inverted generational distance (IGD) is adopted to assess the performance of the algorithms in terms of the convergence and diversity of the obtained non-dominated solutions. The PlatEMO toolbox \citep{tian2017platemo} is used to calculate values of IGD in our experiments. Let $P^{\ast}$ be a set of uniformly distributed solutions sampled from the objective space along the true PF, and $P$ be an obtained approximation to the PF. $IGD$ measures the IGD from $P^{\ast}$ to $P$ as follows:
\begin{equation}
IGD(P^{\ast },P)=\frac{\sum_{\upsilon \in P^{\ast } }d(\upsilon,P)}{|P^{\ast }|}
\end{equation}
where $d(\upsilon,P)$ is the minimum Euclidean distance between $\upsilon$ and all points in $P$. The smaller IGD value, the better the achieved solution set is.
\subsection{Experimental Settings}
In order to test the effectiveness of the proposed TL scheme, two state-of-the-art SAEAs, HK-RVEA \citep{chugh2018surrogate} and T-SAEA \citep{wang2020TSAEA}, and a representative non-surrogate based method including four algorithm schemes in \citep{allmendinger2015multiobjective} are adopted for comparison with TC-SAEA. Specifically, HK-RVEA and T-SAEA are GP-based EA for solving bi-objective optimization problems under consideration; \emph{Waiting}, \emph{Fast-first}, \emph{Brood interleaving} (BI) and \emph{Speculative interleaving} (SI) in \citep{allmendinger2015multiobjective} are four distinct EAs without surrogates. We also include K-RVEA \citep{chugh2018hkrvea} for comparison, which is a GP-based SAEA and proposed for solving MOPs in undelayed environment. Here, we adopt RVEA as the multi-objective optimizer and the standard genetic algorithm (GA) as the single-objective optimizer in all the compared algorithms for the sake of fairness.

TC-SAEA and its variants are implemented in MATLAB R2019a on an Intel Core i7 with 2.21 GHz CPU, and the code of the compared algorithms is provided by their authors. The GP model is constructed using the DACE toolbox \citep{lophaven2002dace}. The general parameter settings in the experiments are given as follows:
\begin{itemize}
\item The number of initial training points $N_{train}=100$.
\item The maximum number of generations before updating GPs $w_{max}=20$.
\item The number of new samples for updating GPs .
\item $\tau$ is set as 5 and 10, respectively, to investigate the impact of the ratio between the two evaluation times.
\item The maximum number of slow objective function evaluations $FE_{s}^{max}=200$.
\end{itemize}

\subsection{Comparison with Some State-of-the-art algorithms \label{result1}}

\begin{table}[thp]
\center
\caption{Statistical results of the IGD values obtained by Waiting, Fast-first, BI, SI, K-RVEA, HK-RVEA, T-SAEA and TC-SAEA with $FE_{s}^{max}=200$ and $\tau=5$}
\label{Tab.1}
\setlength{\tabcolsep}{0.5mm}{
\begin{tabular}{l|ccc|ccc|ccc|ccc|ccc|ccc|ccc|cc}
\hline
\multicolumn{1}{c|}{\multirow{2}{*}{Problem}} & \multicolumn{3}{c|}{Waiting} & \multicolumn{3}{c|}{Fast-first} &
\multicolumn{3}{c|}{BI}&
\multicolumn{3}{c|}{SI} & \multicolumn{3}{c|}{K-RVEA} & \multicolumn{3}{c|}{HK-RVEA} &
\multicolumn{3}{c|}{T-SAEA} &
\multicolumn{2}{c}{TC-SAEA} \\
\multicolumn{1}{c|}{} & \multicolumn{2}{c}{mean} & std & \multicolumn{2}{c}{mean} & std &
\multicolumn{2}{c}{mean} & std &
\multicolumn{2}{c}{mean} & std &
\multicolumn{2}{c}{mean} & std & \multicolumn{2}{c}{mean} & std & \multicolumn{2}{c}{mean} & std & mean & std \\ \hline
DTLZ1  & 30.2 & +         & 14.3 & 69.7 & + & 24.1 & 28.6 & + & 10.2 & 48.1 & + & 10.5 & 29.8 & +         & 17.8 & 42.2          & +         & 10.5 & 21.7          & +         & 11.9 & \textbf{20.1} & 8.16  \\
DTLZ1a & 14.2 & +         & 8.48 & 2.62 & + & 0.25 & 15.6 & + & 6.54 & 28.6 & + & 8.22 & 1.03 & +         & 0.32 & 0.52          & +         & 0.18 & 1.06          & +         & 1.00 & \textbf{0.36} & 0.04  \\
DTLZ2  & 0.24 & +         & 0.05 & 0.80 & + & 0.08 & 0.36 & + & 0.05 & 0.38 & + & 0.03 & 0.13 & +         & 0.06 & 0.10          & +         & 0.02 & 0.05          & $\approx$       & 0.03 & \textbf{0.02} & 0.00  \\
DTLZ3  & 349  & +         & 83.4 & 549  & + & 142  & 357  & + & 74.6 & 462  & + & 67.5 & 385  & +         & 59.4 & 354           & +         & 41.9 & 203           & $\approx$ & 100  & \textbf{132}  & 79.28 \\
DTLZ3a & 227  & +         & 75.4 & 546  & + & 86.9 & 313  & + & 83.1 & 406  & + & 96.6 & 73.7 & +         & 29.2 & 14.9          & +         & 5.27 & 5.34          & +         & 37.5 & \textbf{2.30} & 0.66  \\
DTLZ4  & 0.51 & +         & 0.32 & 0.78 & + & 0.11 & 0.54 & + & 0.06 & 0.65 & + & 0.10 & 0.45 & $\approx$ & 0.23 & \textbf{0.23} & --        & 0.11 & 0.60          & +         & 0.13 & 0.44          & 0.13  \\
DTLZ5  & 0.27 & +         & 0.06 & 0.86 & + & 0.10 & 0.35 & + & 0.04 & 0.39 & + & 0.03 & 0.14 & +         & 0.05 & 0.09          & +         & 0.02 & 0.05          & $\approx$ & 0.02 & \textbf{0.03} & 0.00  \\
DTLZ6  & 7.31 & +         & 0.52 & 8.79 & + & 0.11 & 7.63 & + & 0.44 & 8.26 & + & 0.13 & 5.14 & +         & 0.77 & 4.10          & +         & 0.54 & \textbf{2.56}          & $\approx$ & 1.21 & 2.62 & 1.95  \\
DTLZ7  & 4.41 & +         & 0.62 & 7.53 & + & 0.39 & 5.53 & + & 0.47 & 5.57 & + & 0.68 & 5.54 & +         & 0.47 & 0.06          & $\approx$ & 0.05 & 1.15          & +         & 0.91 & \textbf{0.05} & 0.08  \\
UF1    & 1.01 & +         & 0.14 & 0.49 & + & 0.04 & 0.36 & + & 0.02 & 0.42 & + & 0.05 & 1.20 & +         & 0.11 & 0.23          & +         & 0.02 & \textbf{0.19} & $\approx$ & 0.02 & \textbf{0.19} & 0.02  \\
UF2    & 0.50 & +         & 0.07 & 0.58 & + & 0.09 & 0.45 & + & 0.03 & 0.51 & + & 0.03 & 0.58 & +         & 0.05 & 0.15          & $\approx$ & 0.02 & 0.14          & $\approx$ & 0.02 & \textbf{0.13} & 0.02  \\
UF3    & 0.97 & +         & 0.08 & 1.22 & + & 0.06 & 0.96 & + & 0.04 & 1.08 & + & 0.07 & 1.10 & +         & 0.06 & 0.54          & +         & 0.05 & \textbf{0.19} & --        & 0.08 & 0.42          & 0.03  \\
UF4    & 0.21 & $\approx$ & 0.01 & 0.24 & + & 0.02 & 0.23 & + & 0.00 & 0.23 & + & 0.01 & 0.19 & $\approx$ & 0.00 & 0.22          & $\approx$ & 0.00 & 0.23          & +         & 0.02 & \textbf{0.19} & 0.01  \\
UF5    & 4.75 & +         & 0.42 & 3.53 & + & 0.26 & 2.84 & + & 0.18 & 3.25 & + & 0.16 & 5.05 & +         & 0.96 & 2.46          & +         & 0.43 & 2.49          & +         & 0.44 & \textbf{2.42} & 0.38  \\
UF6    & 4.36 & +         & 0.63 & 2.28 & + & 0.24 & 1.69 & + & 0.13 & 1.99 & + & 0.17 & 5.26 & +         & 0.59 & 1.34          & +         & 0.13 & 1.01          & +         & 0.25 & \textbf{0.81} & 0.19  \\
UF7    & 1.20 & +         & 0.12 & 0.58 & + & 0.06 & 0.38 & + & 0.03 & 0.47 & + & 0.05 & 1.24 & +         & 0.14 & \textbf{0.21} & --        & 0.04 & 0.37          & $\approx$ & 0.06 & 0.33          & 0.05 \\
\hline
\end{tabular}}
\end{table}

\begin{table}[thp]
\center
\caption{Statistical results of the IGD values obtained by Waiting, Fast-first, BI, SI, K-RVEA, HK-RVEA, T-SAEA and TC-SAEA with $FE_{s}^{max}=200$ and $\tau=10$}
\label{Tab.2}
\setlength{\tabcolsep}{0.5mm}{
\begin{tabular}{l|ccc|ccc|ccc|ccc|ccc|ccc|ccc|cc}
\hline
\multicolumn{1}{c|}{\multirow{2}{*}{Problem}} & \multicolumn{3}{c|}{Waiting} & \multicolumn{3}{c|}{Fast-first} &
\multicolumn{3}{c|}{BI}&
\multicolumn{3}{c|}{SI} & \multicolumn{3}{c|}{K-RVEA} & \multicolumn{3}{c|}{HK-RVEA} &
\multicolumn{3}{c|}{T-SAEA} &
\multicolumn{2}{c}{TC-SAEA} \\
\multicolumn{1}{c|}{} & \multicolumn{2}{c}{mean} & std & \multicolumn{2}{c}{mean} & std &
\multicolumn{2}{c}{mean} & std &
\multicolumn{2}{c}{mean} & std &
\multicolumn{2}{c}{mean} & std & \multicolumn{2}{c}{mean} & std & \multicolumn{2}{c}{mean} & std & mean & std \\ \hline
DTLZ1  & 30.2 & +         & 14.3 & 102  & + & 35.5 & 30.1 & +         & 8.24 & 44.9 & +         & 15.1 & 29.8          & +  & 17.8 & 41.5          & +         & 11.8 & 35.8          & +         & 18.4 & \textbf{16.8} & 13.9 \\
DTLZ1a & 14.2 & +         & 8.48 & 24.3 & + & 10.8 & 13.1 & +         & 7.95 & 34.8 & +         & 12.3 & 1.03          & +  & 0.32 & 1.04          & +         & 0.47 & 0.81          & +         & 0.68 & \textbf{0.60} & 0.40 \\
DTLZ2  & 0.24 & +         & 0.05 & 0.96 & + & 0.11 & 0.35 & +         & 0.04 & 0.39 & +         & 0.04 & 0.13          & +  & 0.06 & 0.08          & +         & 0.01 & 0.06          & $\approx$ & 0.03 & \textbf{0.03} & 0.02 \\
DTLZ3  & 349  & +         & 83.4 & 651  & + & 115  & 369  & +         & 59.6 & 430  & +         & 90.7 & 385           & +  & 59.4 & 379           & +         & 30.5 & 385           & +         & 118  & \textbf{137}  & 77.0 \\
DTLZ3a & 227  & +         & 75.4 & 598  & + & 139  & 284  & +         & 68.7 & 418  & +         & 90.2 & 73.7          & +  & 29.2 & 19.8          & +         & 8.50 & 6.71          & +         & 6.68 & \textbf{0.43} & 0.21 \\
DTLZ4  & 0.51 & +         & 0.32 & 0.75 & + & 0.13 & 0.48 & +         & 0.07 & 0.66 & +         & 0.12 & 0.45          & -- & 0.15 & \textbf{0.17} & --        & 0.10 & 0.47          & $\approx$ & 0.17 & 0.48          & 0.17 \\
DTLZ5  & 0.27 & +         & 0.06 & 0.97 & + & 0.10 & 0.37 & +         & 0.04 & 0.41 & +         & 0.03 & 0.14          & +  & 0.05 & 0.09          & +         & 0.01 & 0.06          & $\approx$ & 0.03 & \textbf{0.03} & 0.01 \\
DTLZ6  & 7.31 & +         & 0.52 & 8.96 & + & 0.19 & 7.71 & +         & 0.32 & 8.28 & +         & 0.16 & 5.14          & +  & 0.77 & 4.23          & +         & 0.35 & 6.49          & +         & 0.47 & \textbf{3.31} & 0.65 \\
DTLZ7  & 4.41 & +         & 0.62 & 7.81 & + & 0.64 & 5.89 & +         & 0.36 & 5.79 & +         & 0.50 & 5.54          & +  & 0.47 & \textbf{0.04} & --        & 0.03 & 4.26          & +         & 0.66 & 0.08          & 0.21 \\
UF1    & 1.01 & +         & 0.14 & 0.54 & + & 0.07 & 0.42 & +         & 0.04 & 0.41 & +         & 0.04 & 1.20          & +  & 0.11 & 0.26          & +         & 0.04 & 0.64          & +         & 0.18 & \textbf{0.15} & 0.02 \\
UF2    & 0.50 & +         & 0.07 & 0.74 & + & 0.10 & 0.44 & +         & 0.03 & 0.52 & +         & 0.04 & 0.58          & +  & 0.05 & 0.15          & +         & 0.01 & 0.27          & +         & 0.08 & \textbf{0.12} & 0.02 \\
UF3    & 0.97 & +         & 0.08 & 1.21 & + & 0.07 & 0.91 & +         & 0.05 & 1.09 & +         & 0.08 & 1.10          & +  & 0.06 & 0.53          & $\approx$ & 0.03 & 0.58          & +         & 0.07 & \textbf{0.51} & 0.06 \\
UF4    & 0.21 & $\approx$ & 0.01 & 0.58 & + & 0.02 & 0.23 & $\approx$ & 0.00 & 0.23 & $\approx$ & 0.01 & \textbf{0.19} & -- & 0.00 & 0.23          & $\approx$ & 0.01 & \textbf{0.19} & --        & 0.01 & 0.22          & 0.00 \\
UF5    & 4.75 & +         & 0.42 & 3.46 & + & 0.28 & 2.96 & +         & 0.14 & 3.24 & +         & 0.19 & 5.05          & +  & 0.31 & 2.43          & $\approx$ & 0.32 & 4.12          & +         & 0.63 & \textbf{2.39} & 0.20 \\
UF6    & 4.36 & +         & 0.63 & 2.17 & + & 0.25 & 1.55 & +         & 0.14 & 2.00 & +         & 0.18 & 5.26          & +  & 0.59 & 1.00          & +        & 0.23 & 1.42          & +       & 0.79 & \textbf{0.69} & 0.12 \\
UF7    & 1.11 & +         & 0.18 & 0.54 & + & 0.06 & 0.37 & +         & 0.03 & 0.46 & +         & 0.05 & 1.24          & +  & 0.14 & \textbf{0.26} & $\approx$ & 0.05 & 0.59          & +         & 0.20 & 0.27          & 0.13 \\
\hline
\end{tabular}}
\end{table}

\begin{figure}[!htb]
\centerline{
\subfloat[]{\includegraphics[width=1.9in]{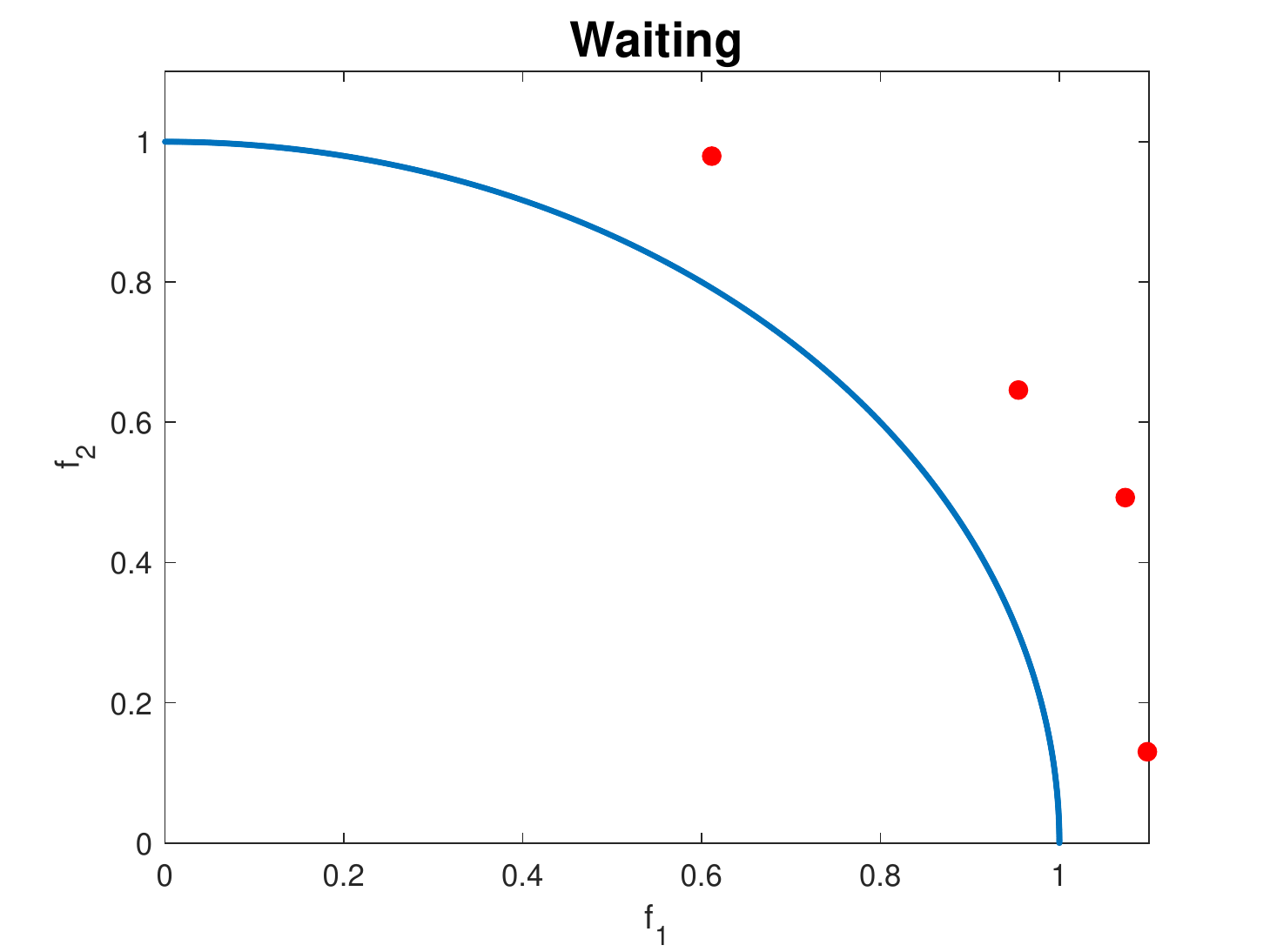}%
\hfil
}
\subfloat[]{\includegraphics[width=1.9in]{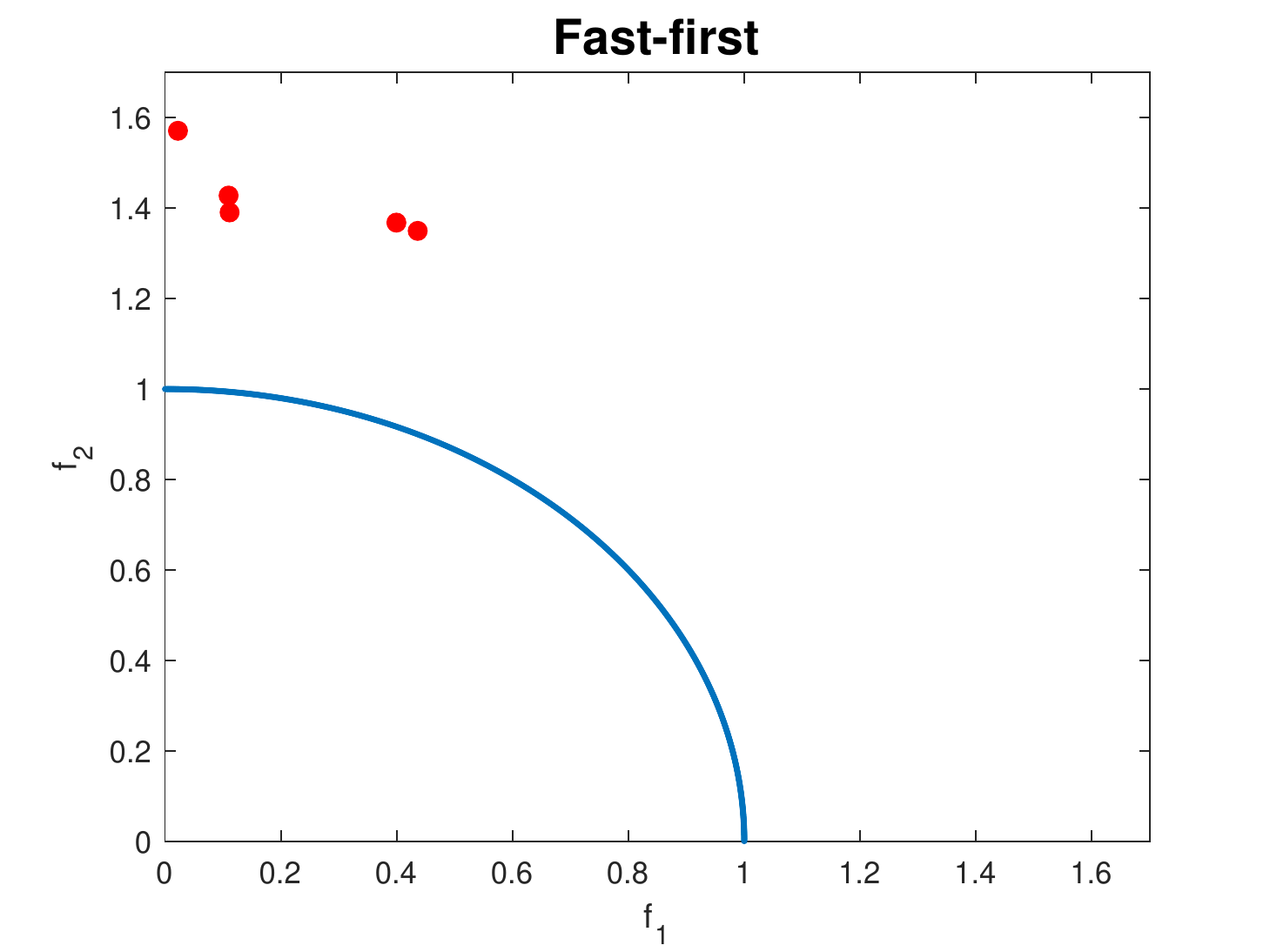}%
\hfil
}
\subfloat[]{\includegraphics[width=1.9in]{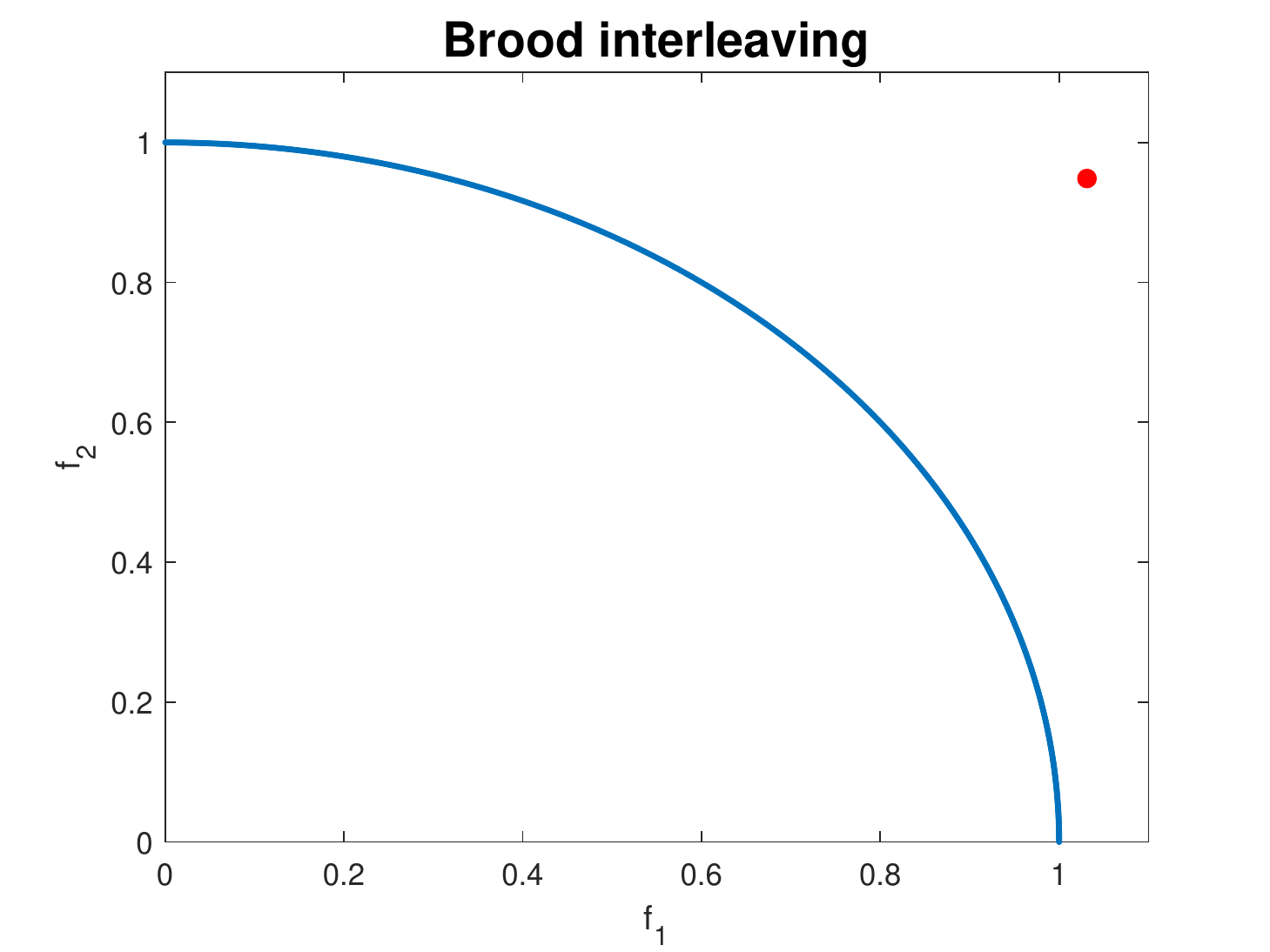}%
\hfil
}
\subfloat[]{\includegraphics[width=1.9in]{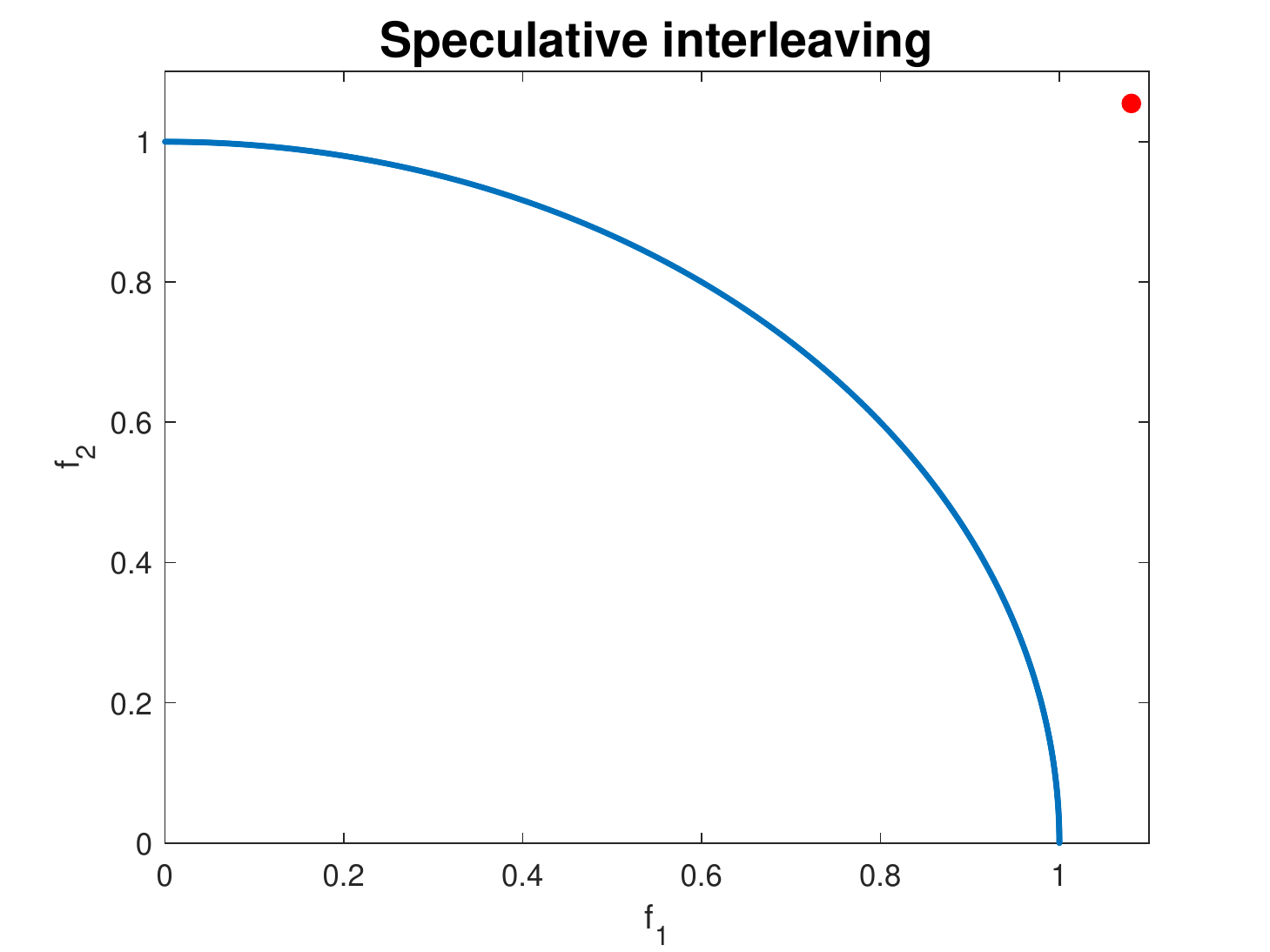}%
\hfil
}}
\centerline{
\subfloat[]{\includegraphics[width=1.9in]{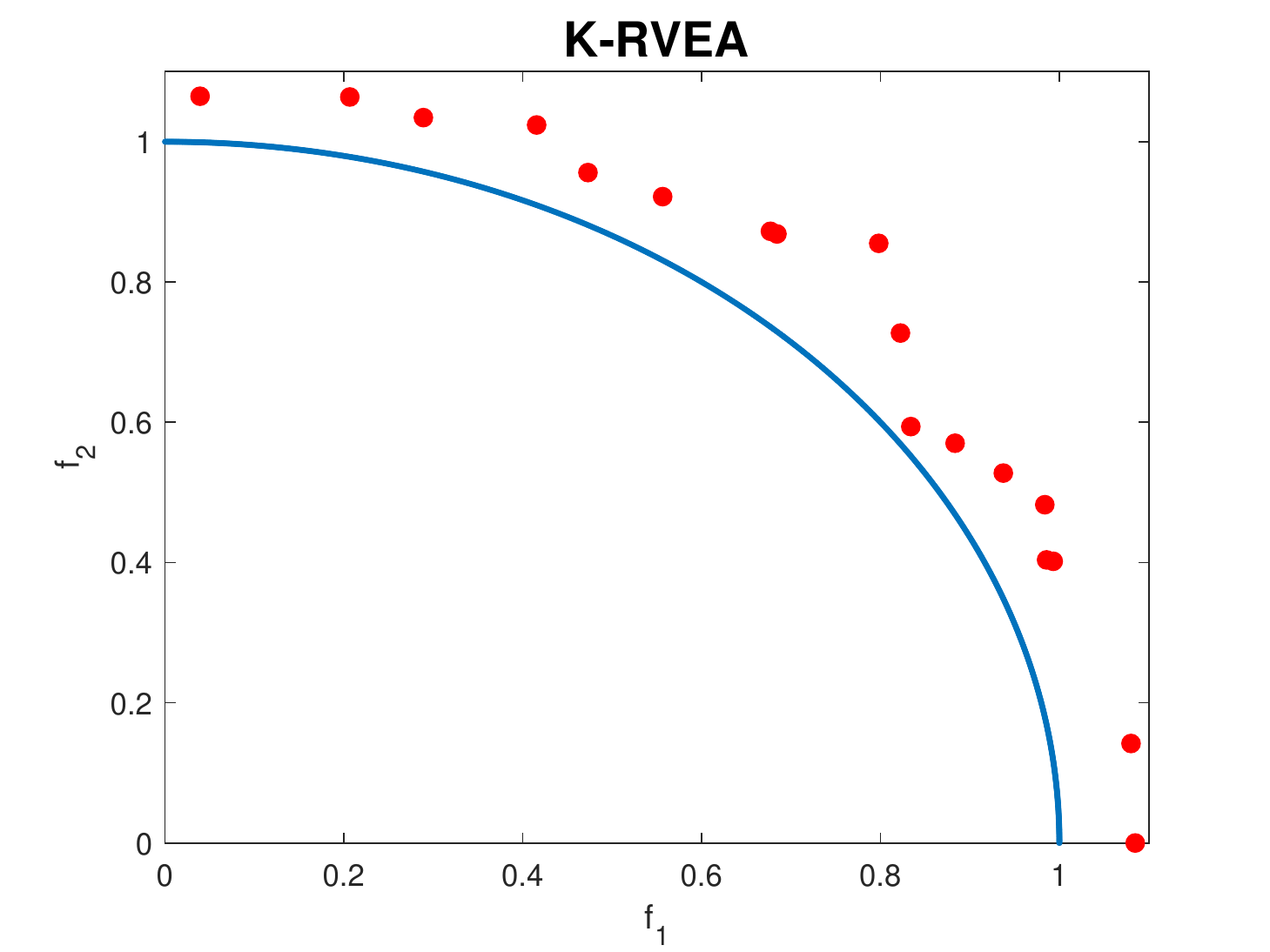}%
\hfil
}
\subfloat[]{\includegraphics[width=1.9in]{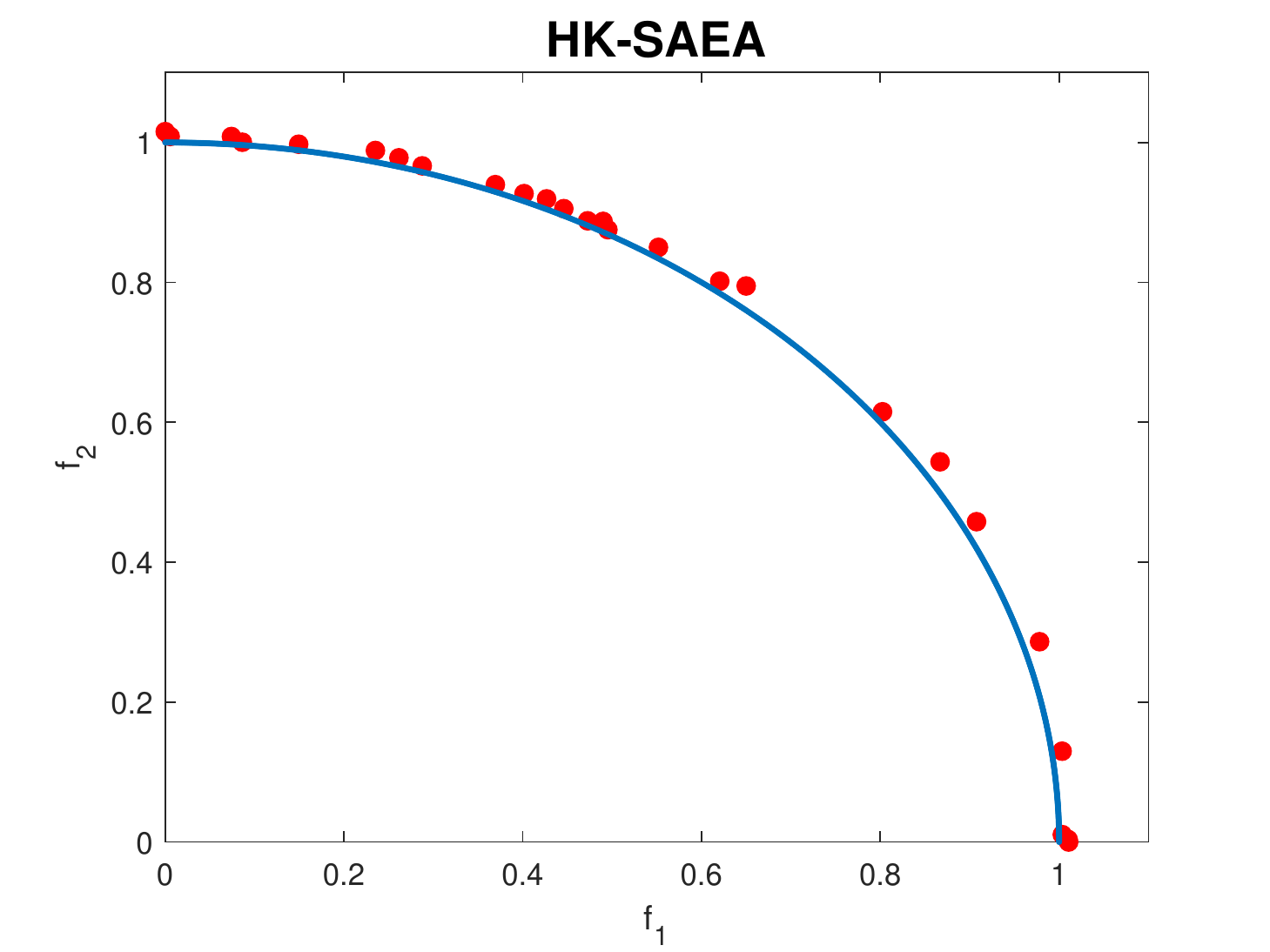}%
\hfil
}
\subfloat[]{\includegraphics[width=1.9in]{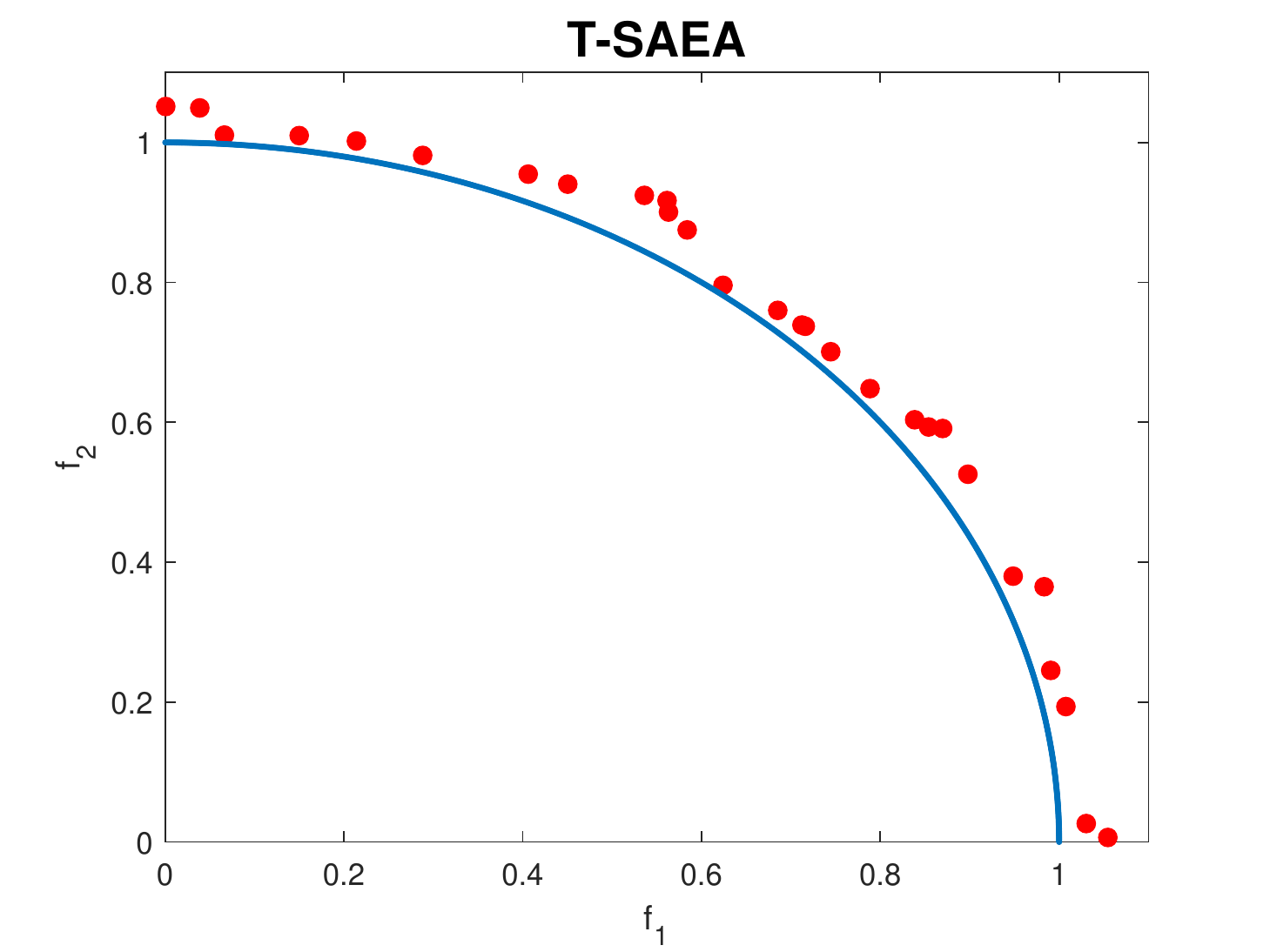}%
\hfil
}
\subfloat[]{\includegraphics[width=1.9in]{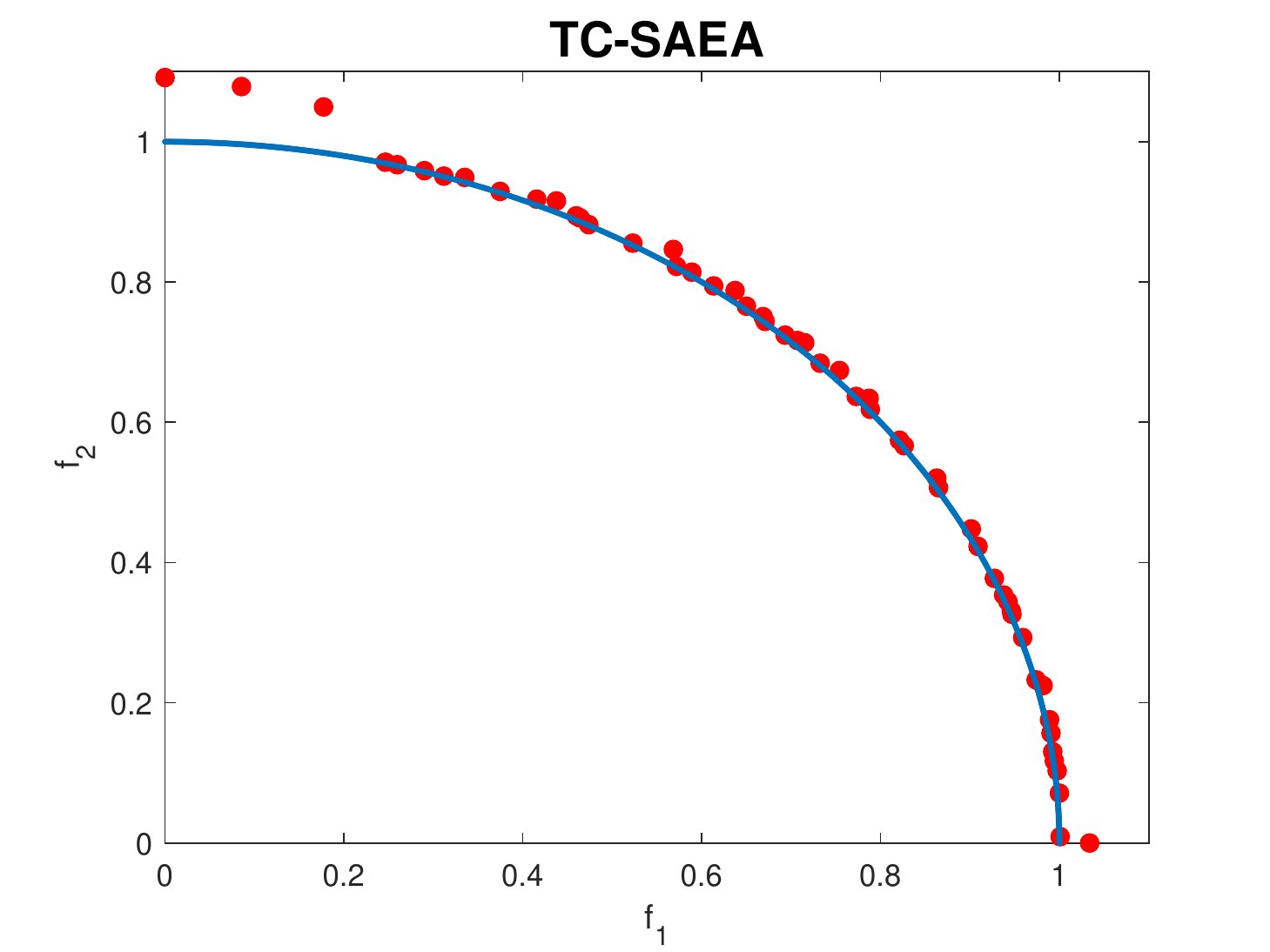}%
\hfil
}}
\caption{The final non-dominated solutions obtained by the compared algorithms on the bi-objective DTLZ2 in the run associated with the median IGD value.}
\label{Fig.4}
\end{figure}

\begin{figure*}[!htb]
\centerline{
\subfloat[]{\includegraphics[width=1.9in]{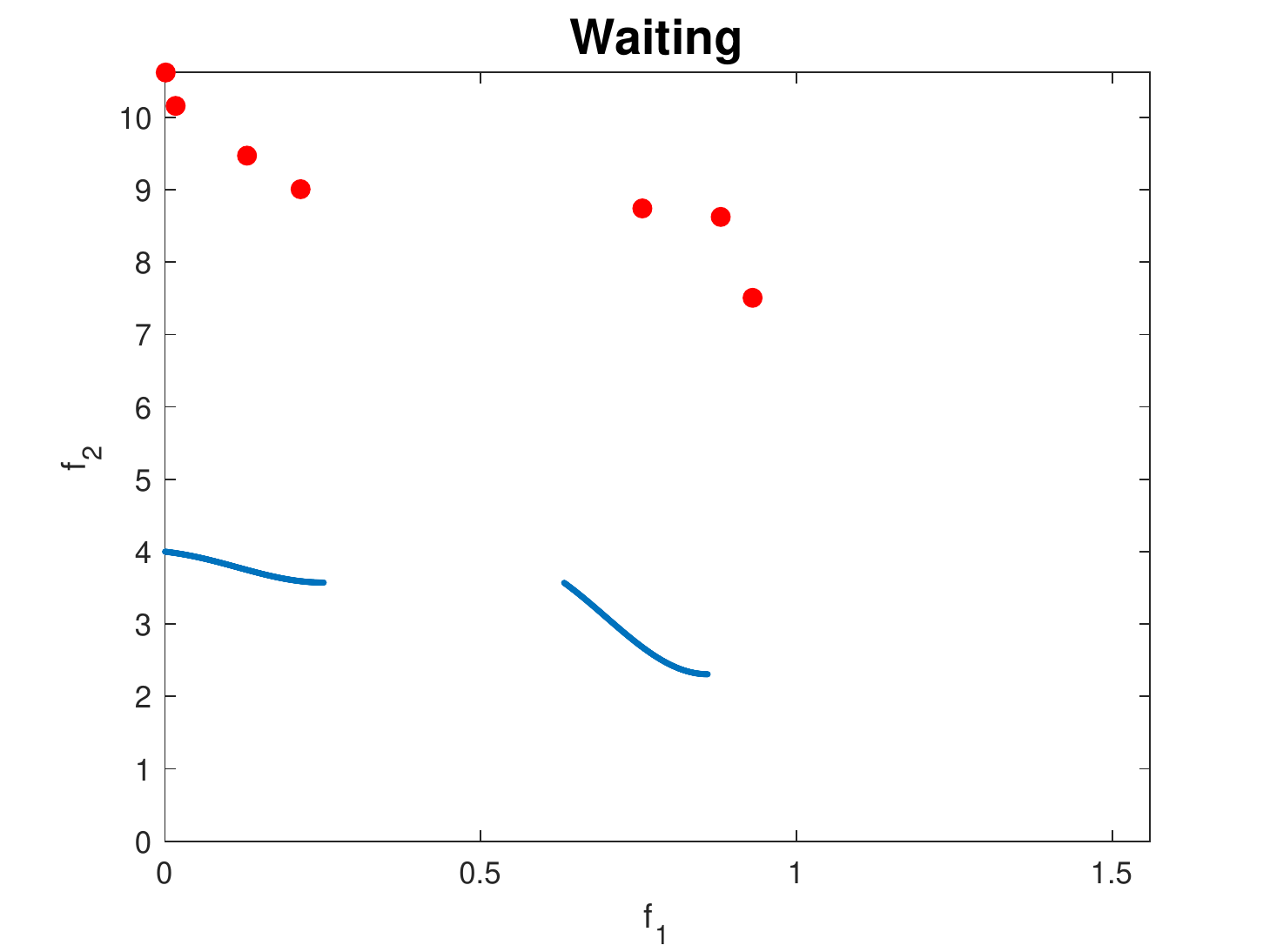}%
\hfil
}
\subfloat[]{\includegraphics[width=1.9in]{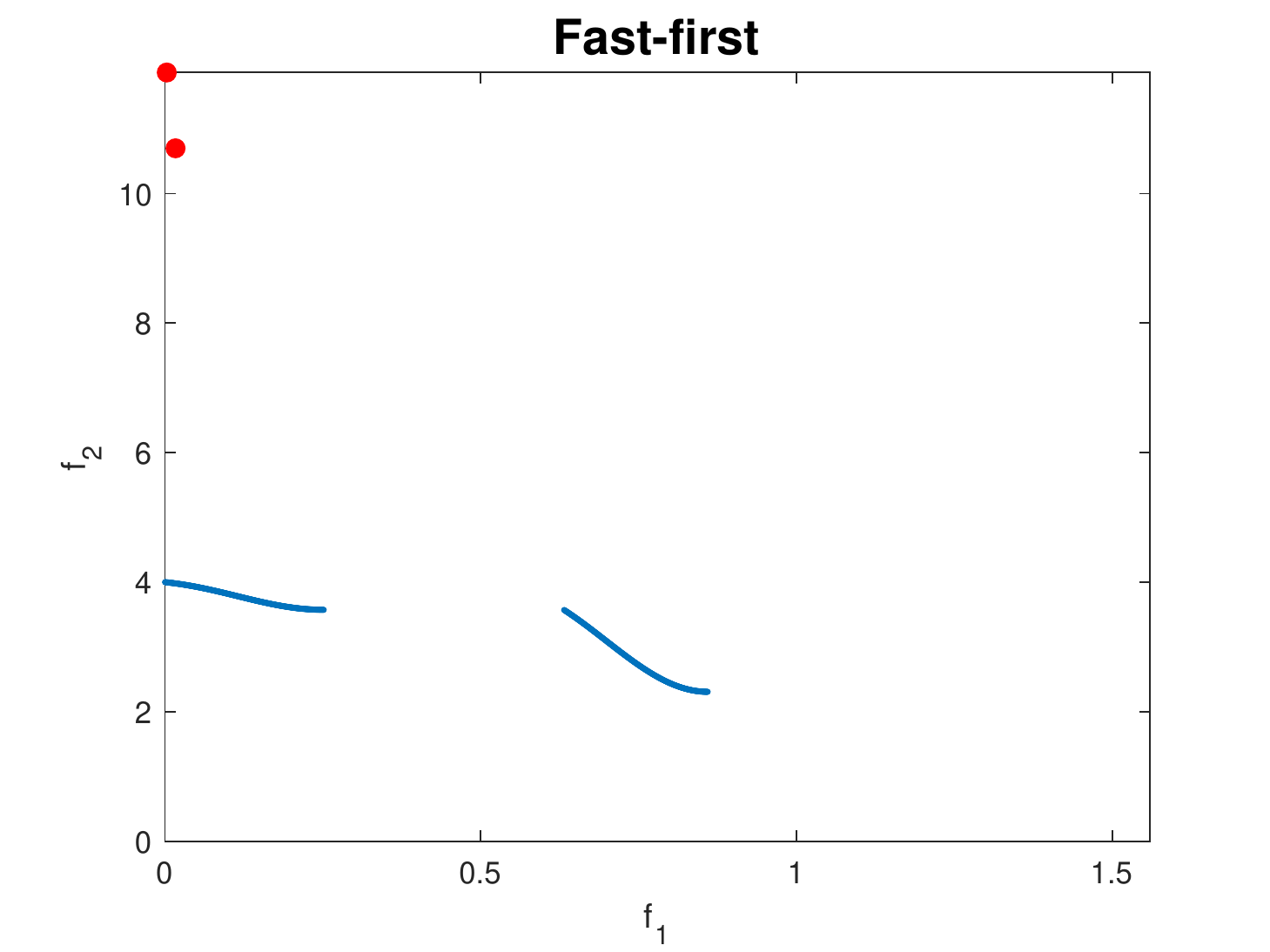}%
\hfil
}
\subfloat[]{\includegraphics[width=1.9in]{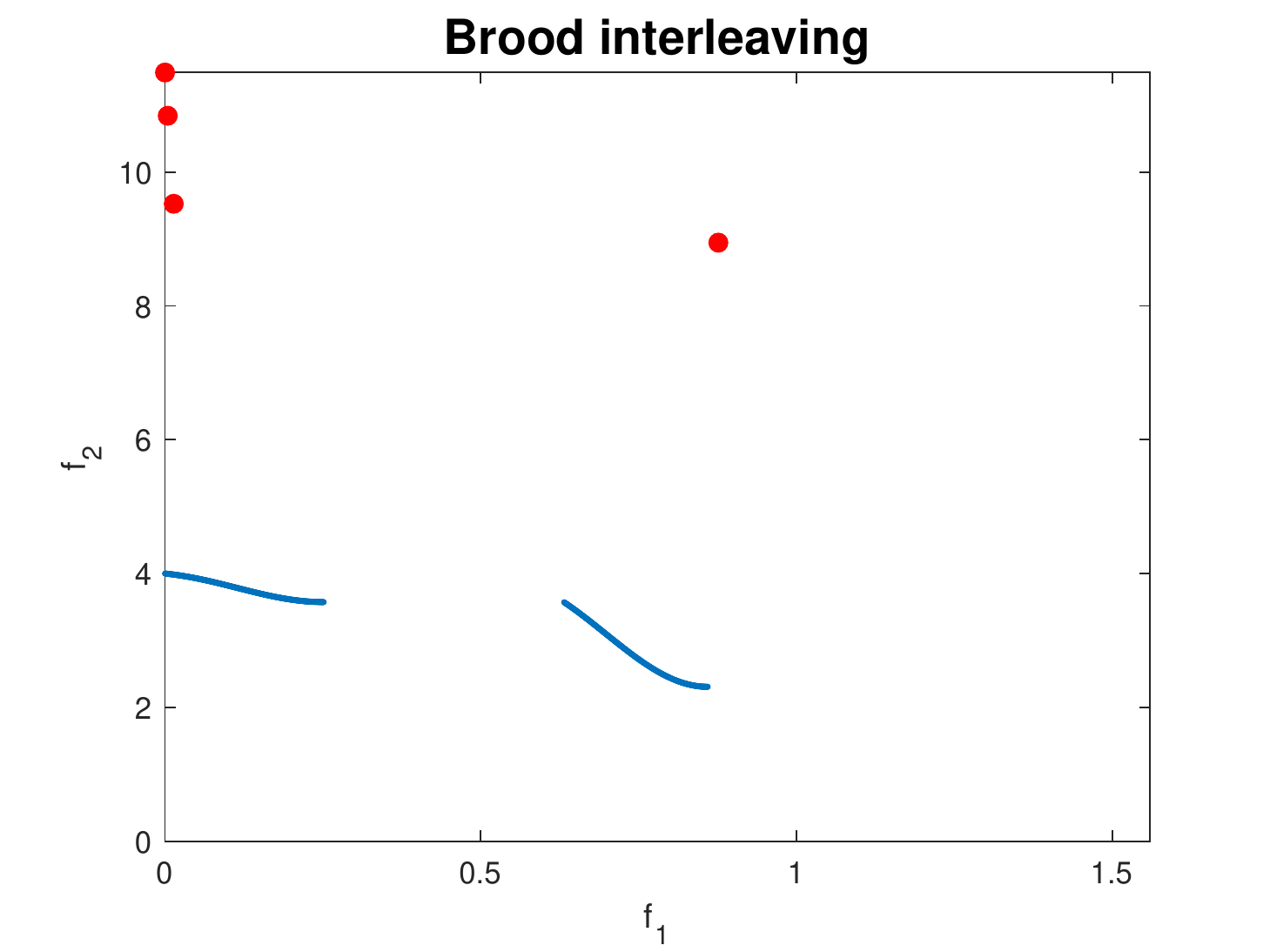}%
\hfil
}
\subfloat[]{\includegraphics[width=1.9in]{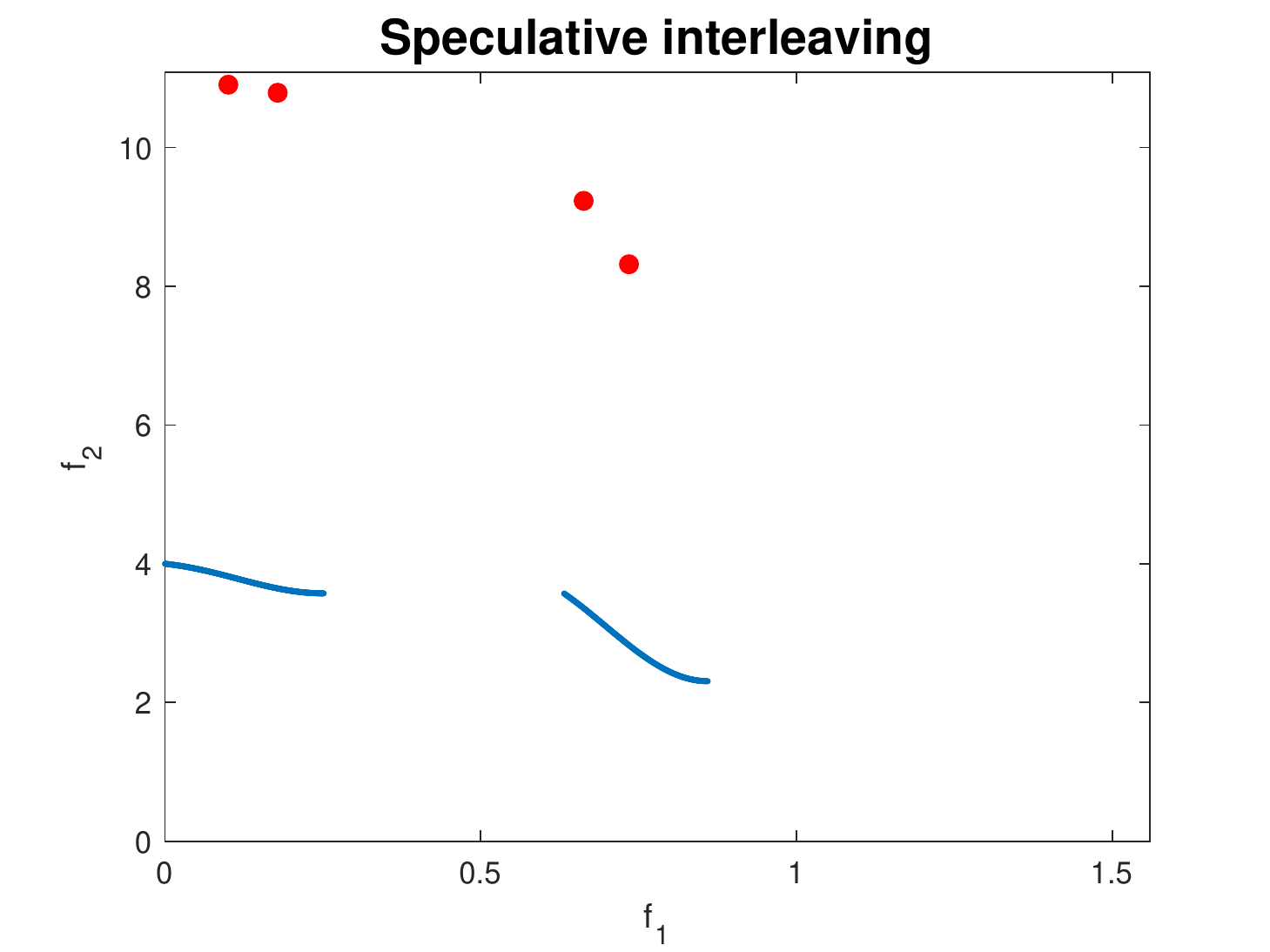}%
\hfil
}
}
\centerline{
\subfloat[]{\includegraphics[width=1.9in]{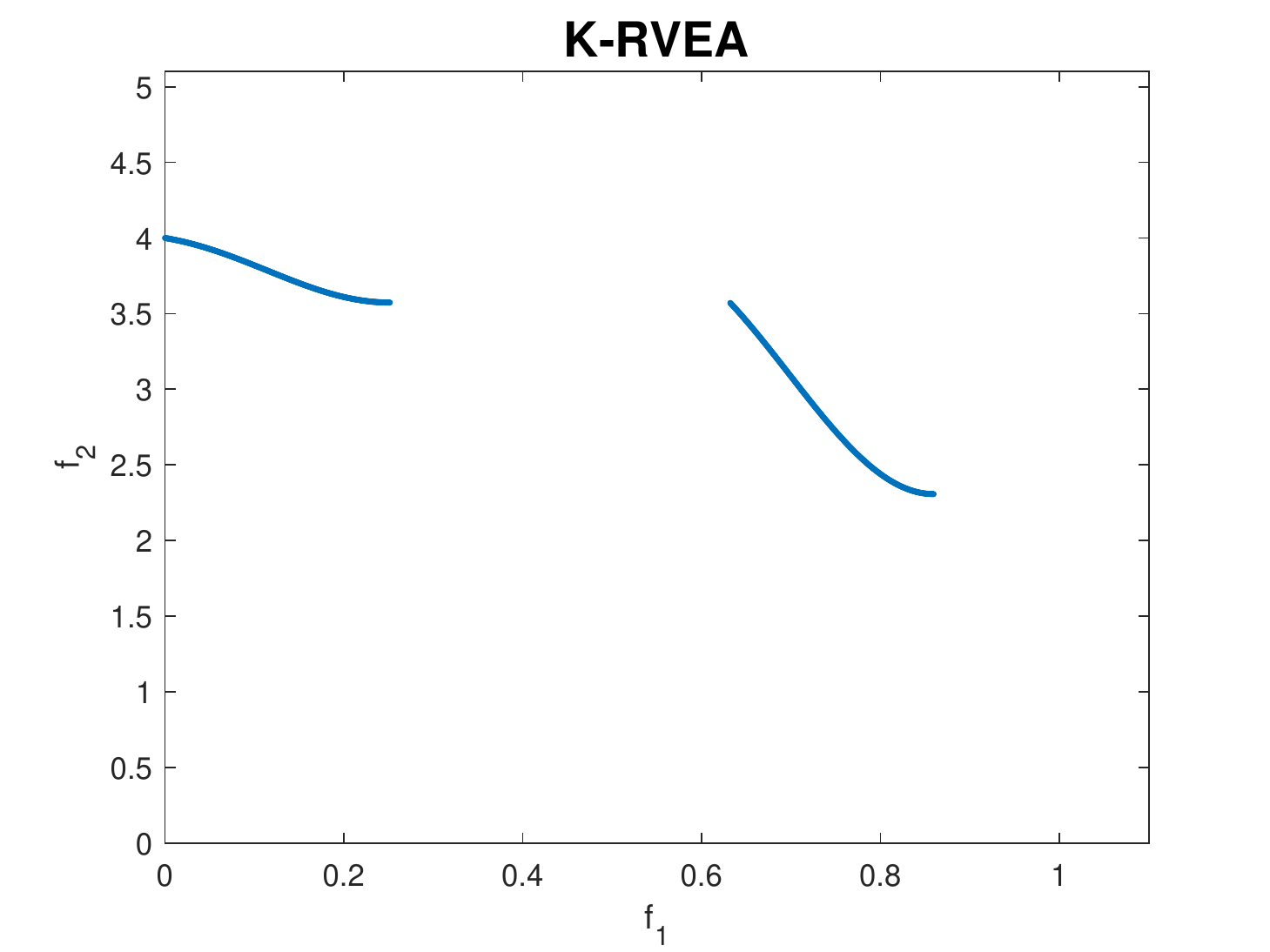}%
\hfil
}
\subfloat[]{\includegraphics[width=1.9in]{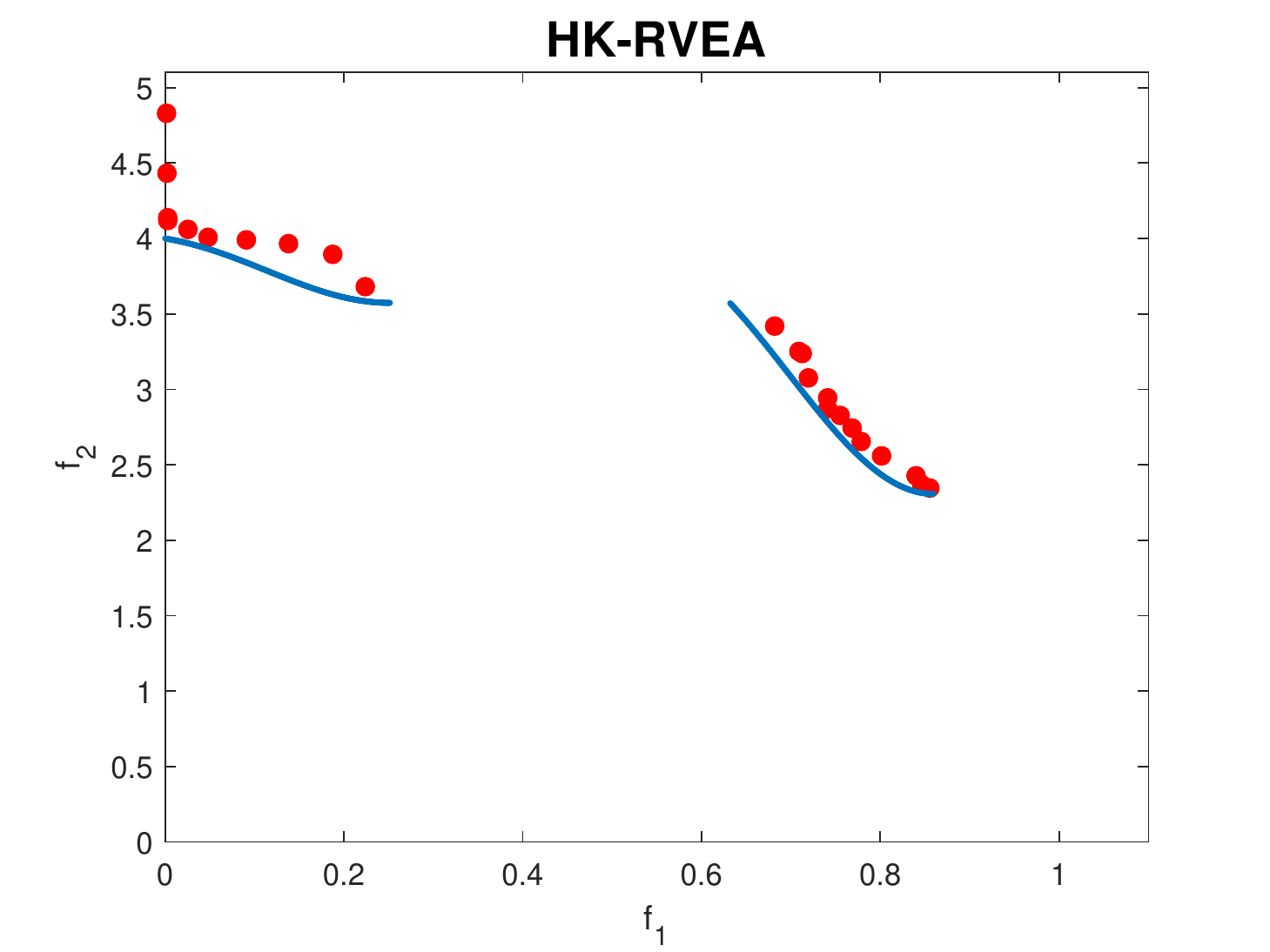}%
\hfil
}
\subfloat[]{\includegraphics[width=1.9in]{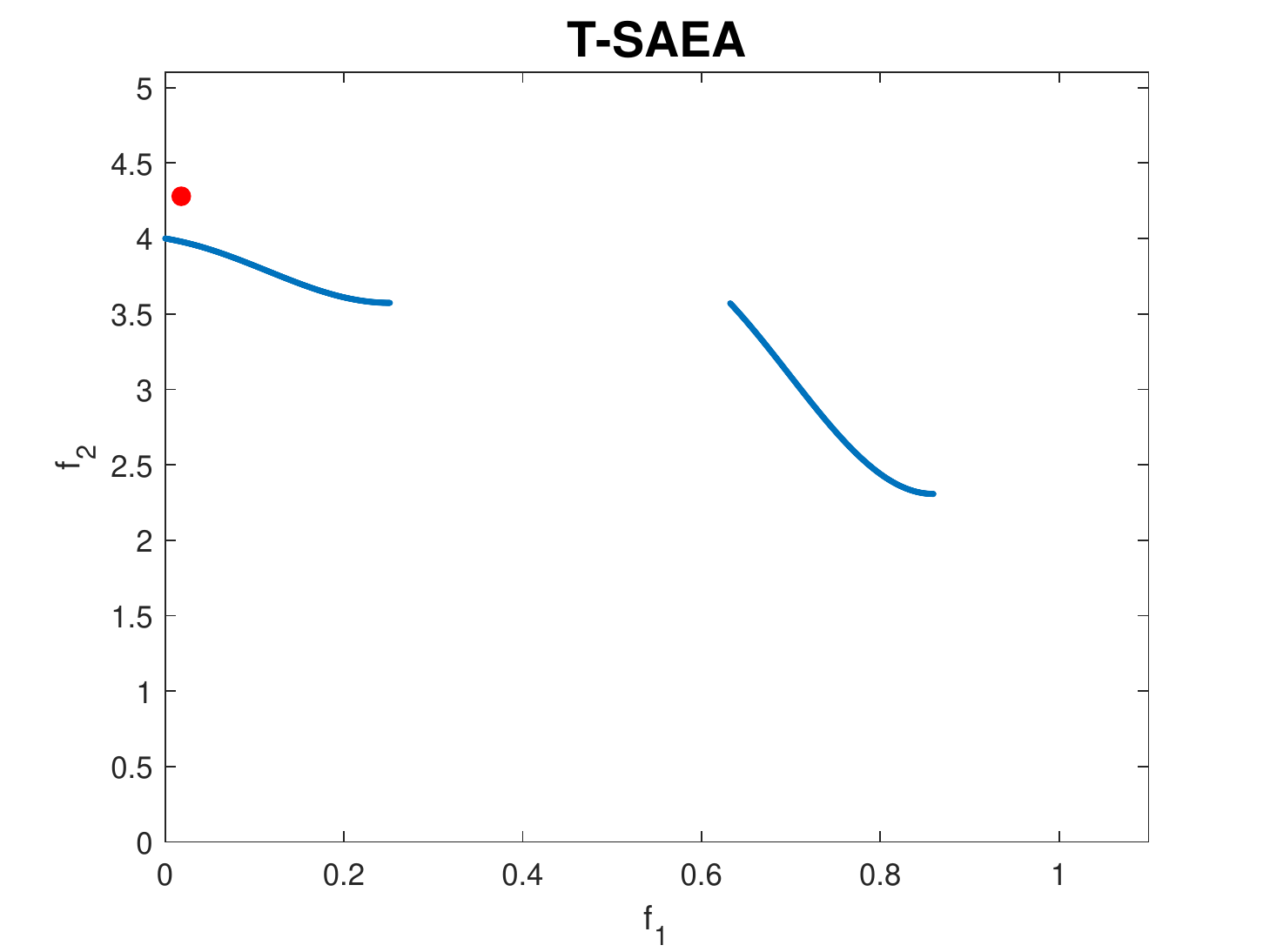}%
\hfil
}
\subfloat[]{\includegraphics[width=1.9in]{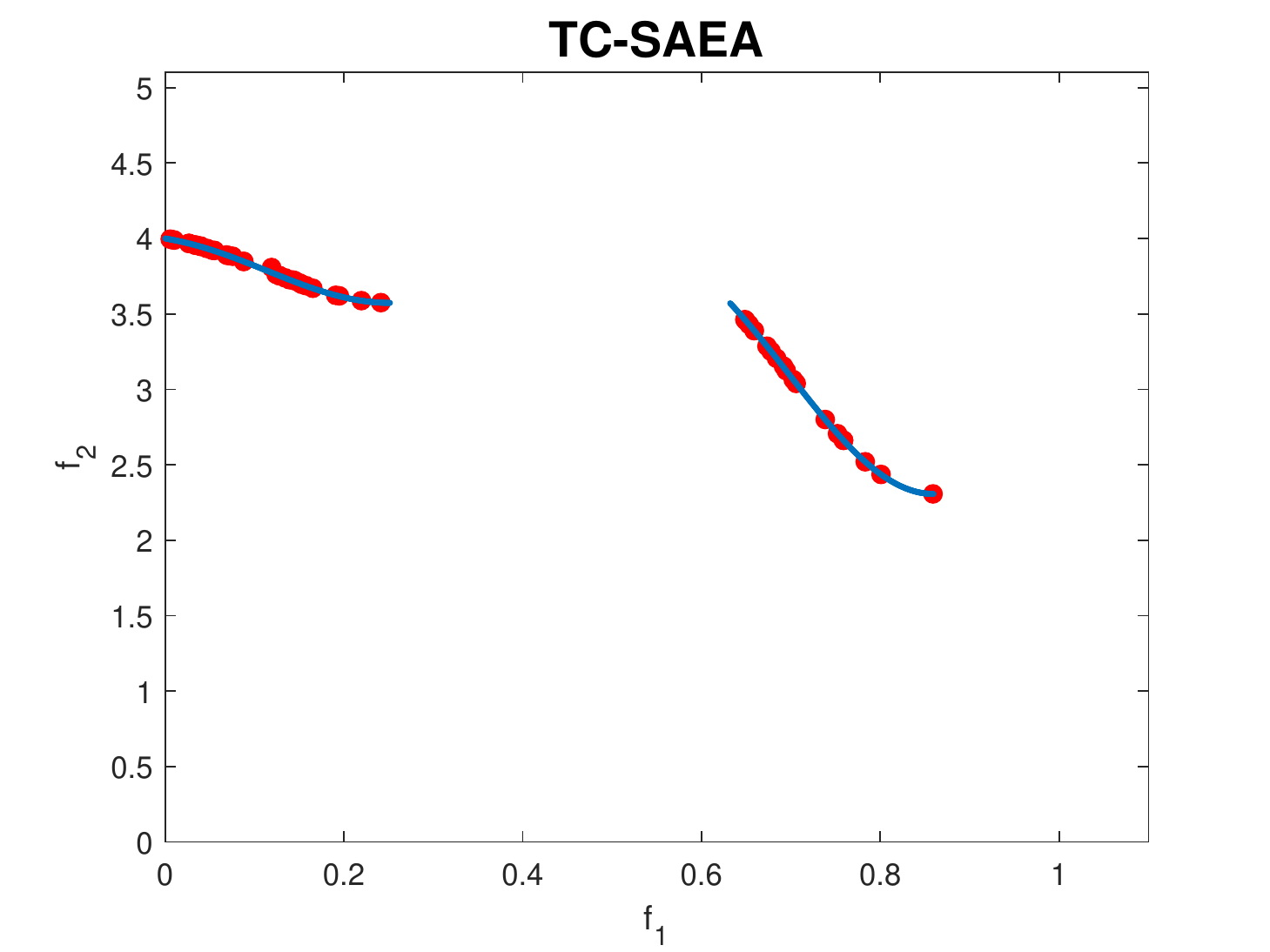}%
\hfil
}
}
\caption{The final non-dominated solutions obtained by the compared algorithms on the bi-objective DTLZ7 in the run associated with the median IGD value.}
\label{Fig.5}
\end{figure*}

\begin{figure*}[!htb]
\centerline{
\subfloat[]{\includegraphics[width=1.9in]{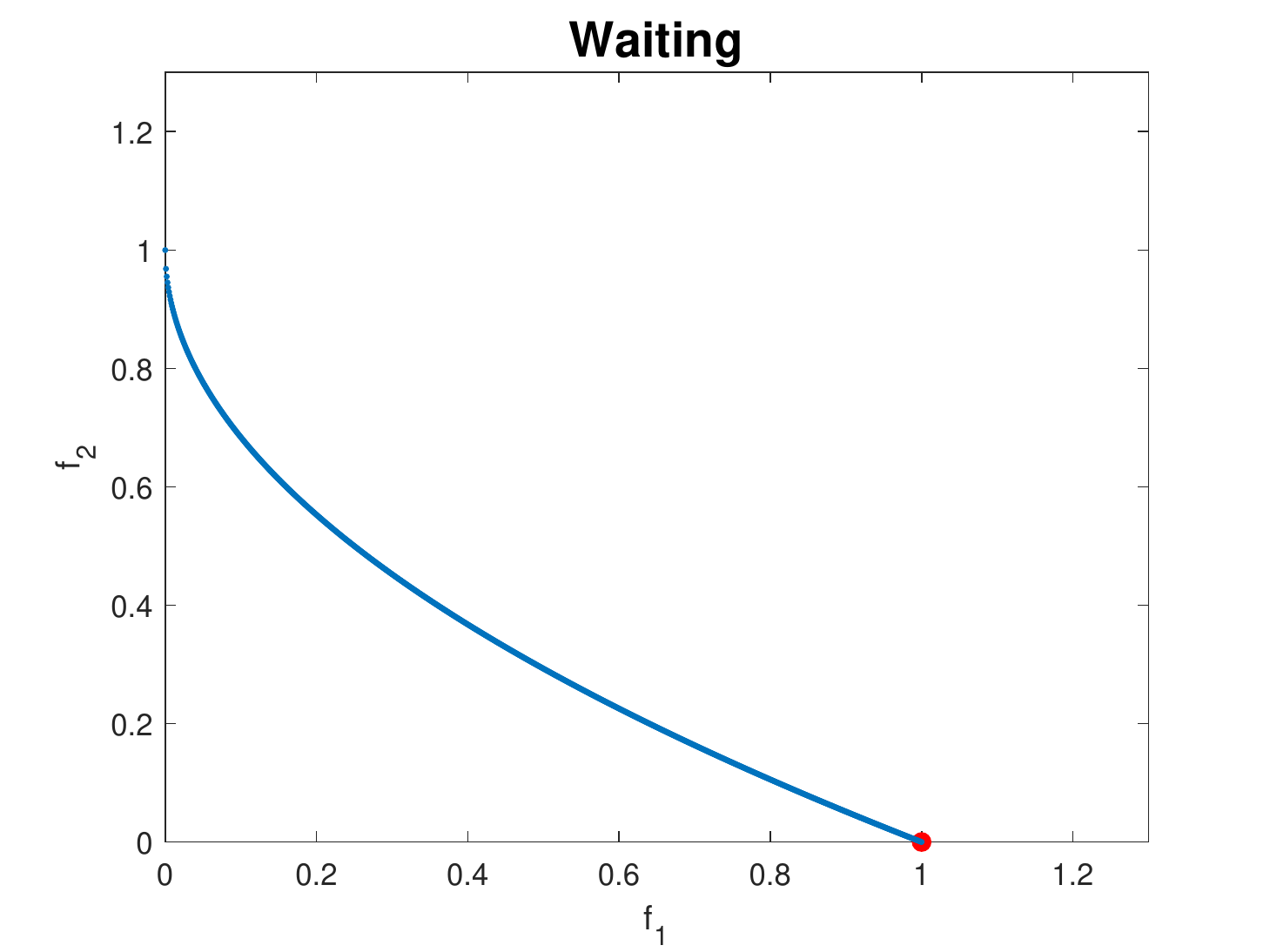}%
\hfil
}
\subfloat[]{\includegraphics[width=1.9in]{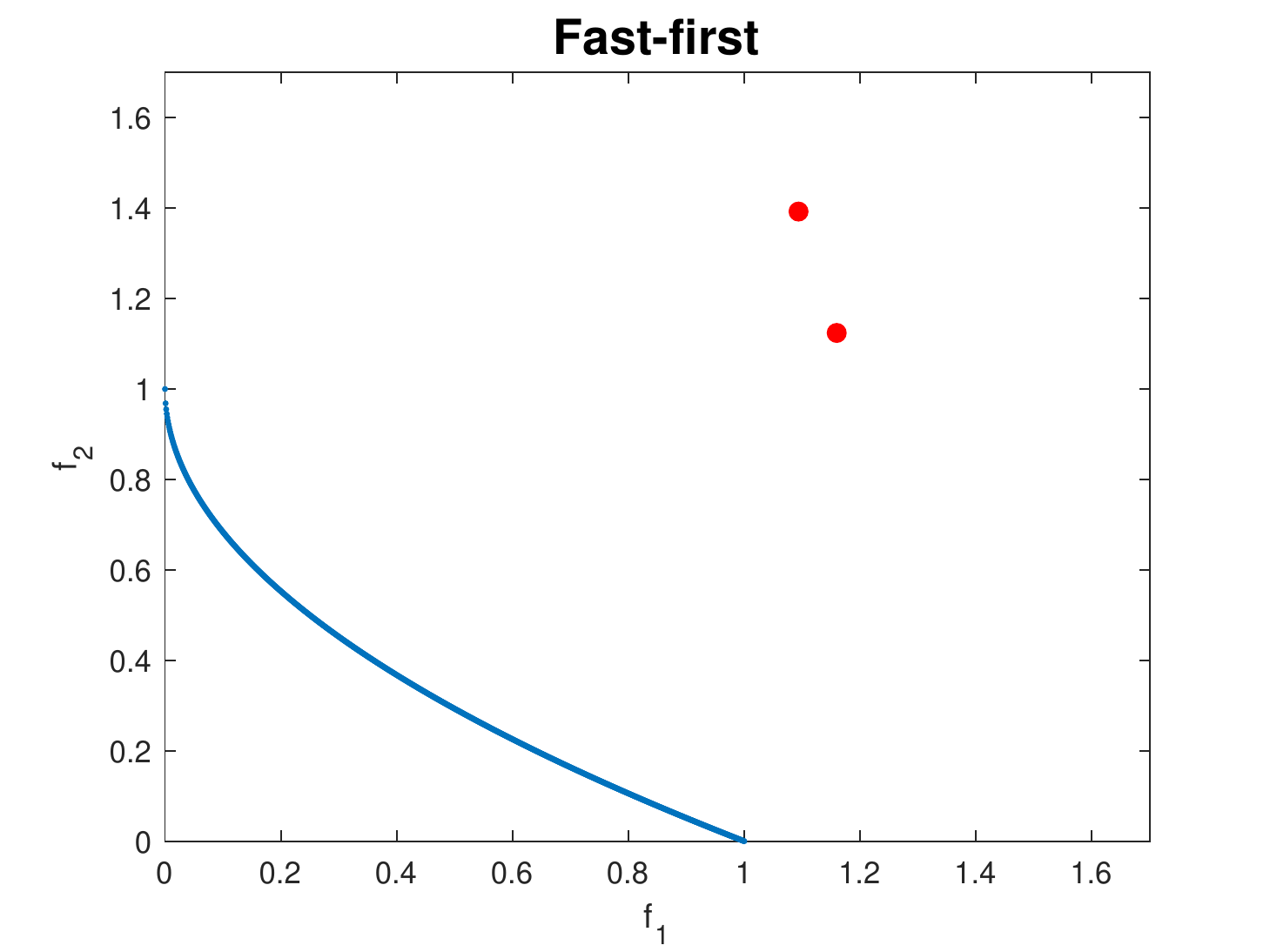}%
\hfil
}
\subfloat[]{\includegraphics[width=1.9in]{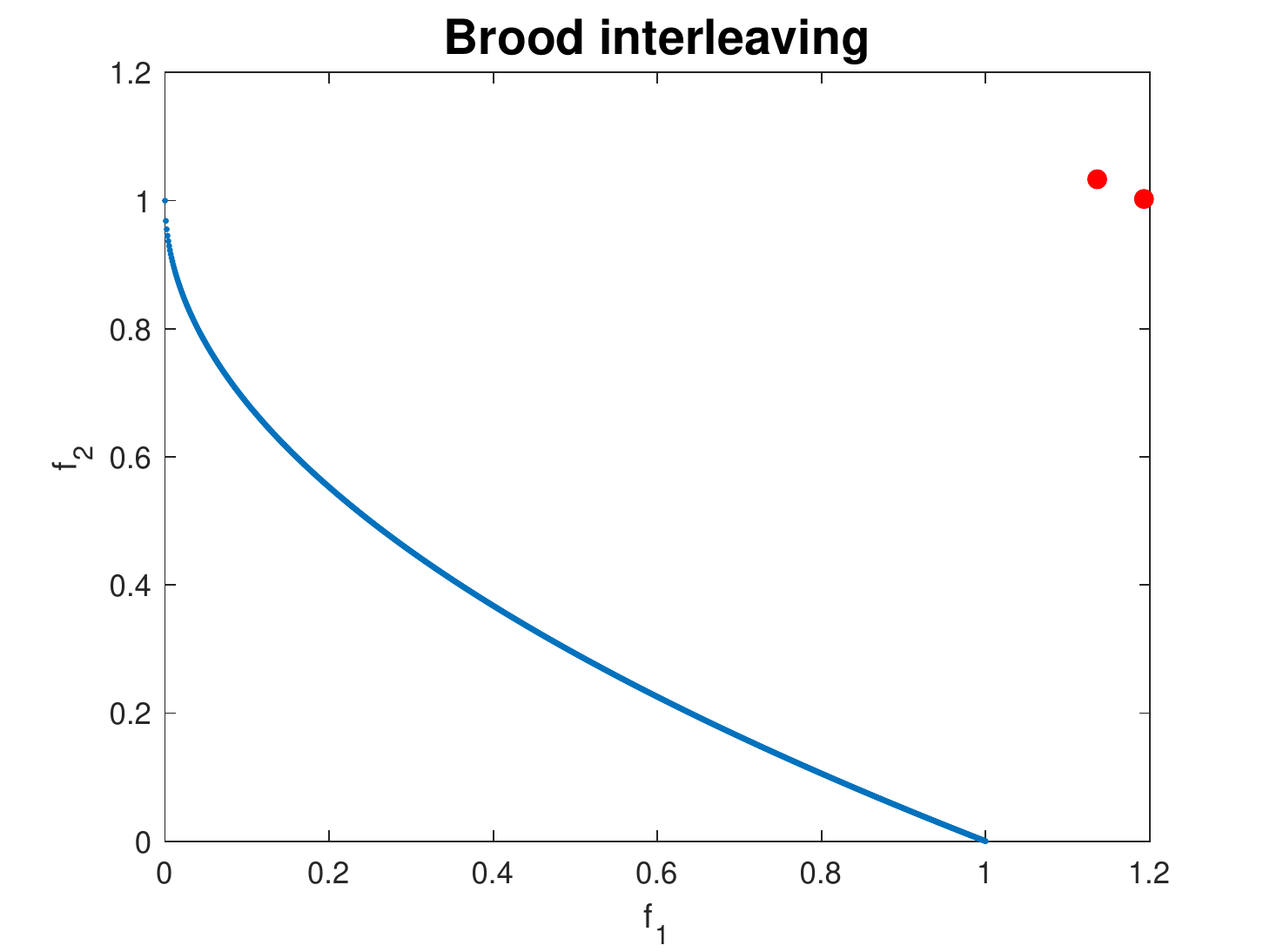}%
\hfil
}
\subfloat[]{\includegraphics[width=1.9in]{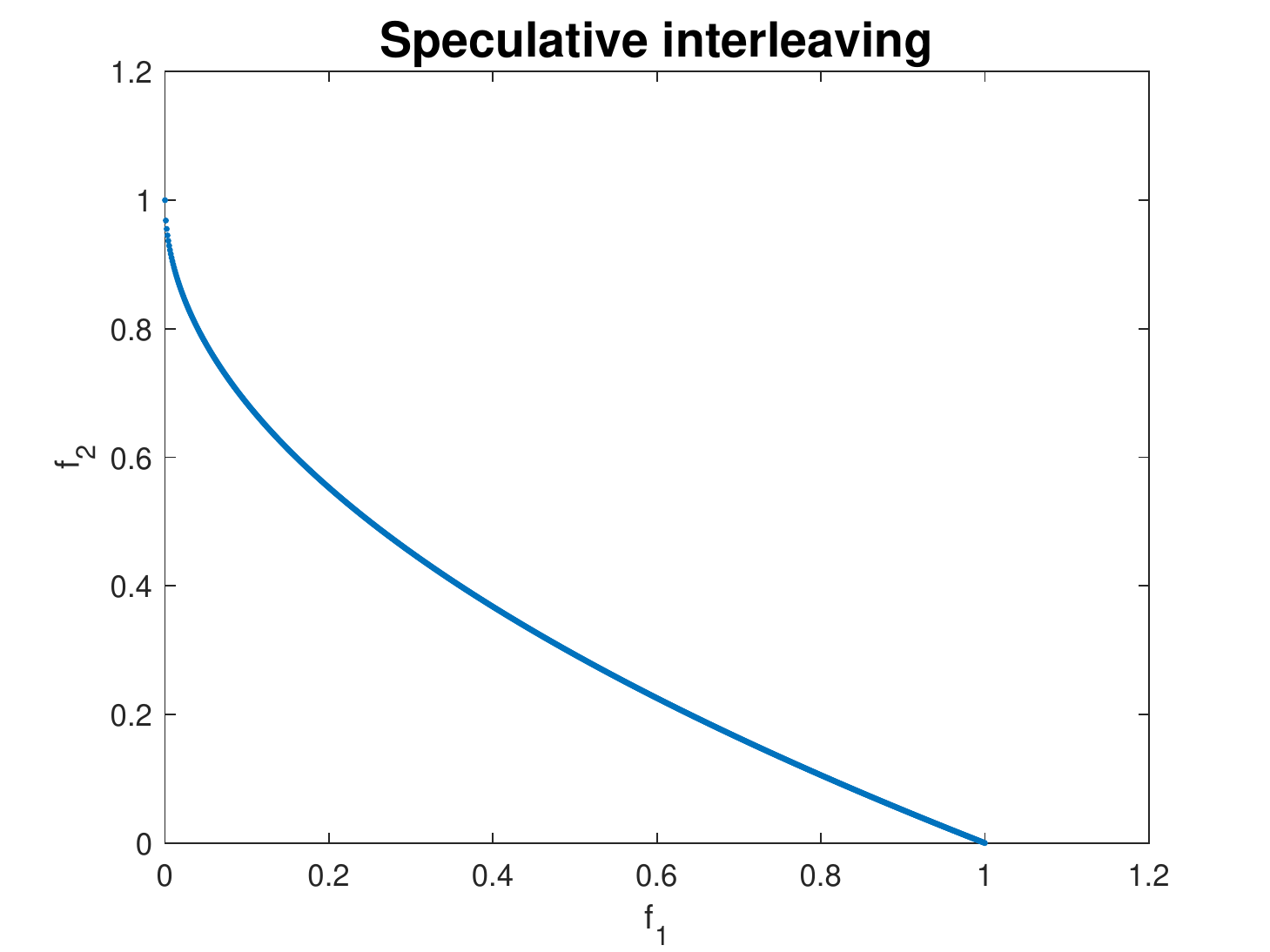}%
\hfil
}
}
\centerline{
\subfloat[]{\includegraphics[width=1.9in]{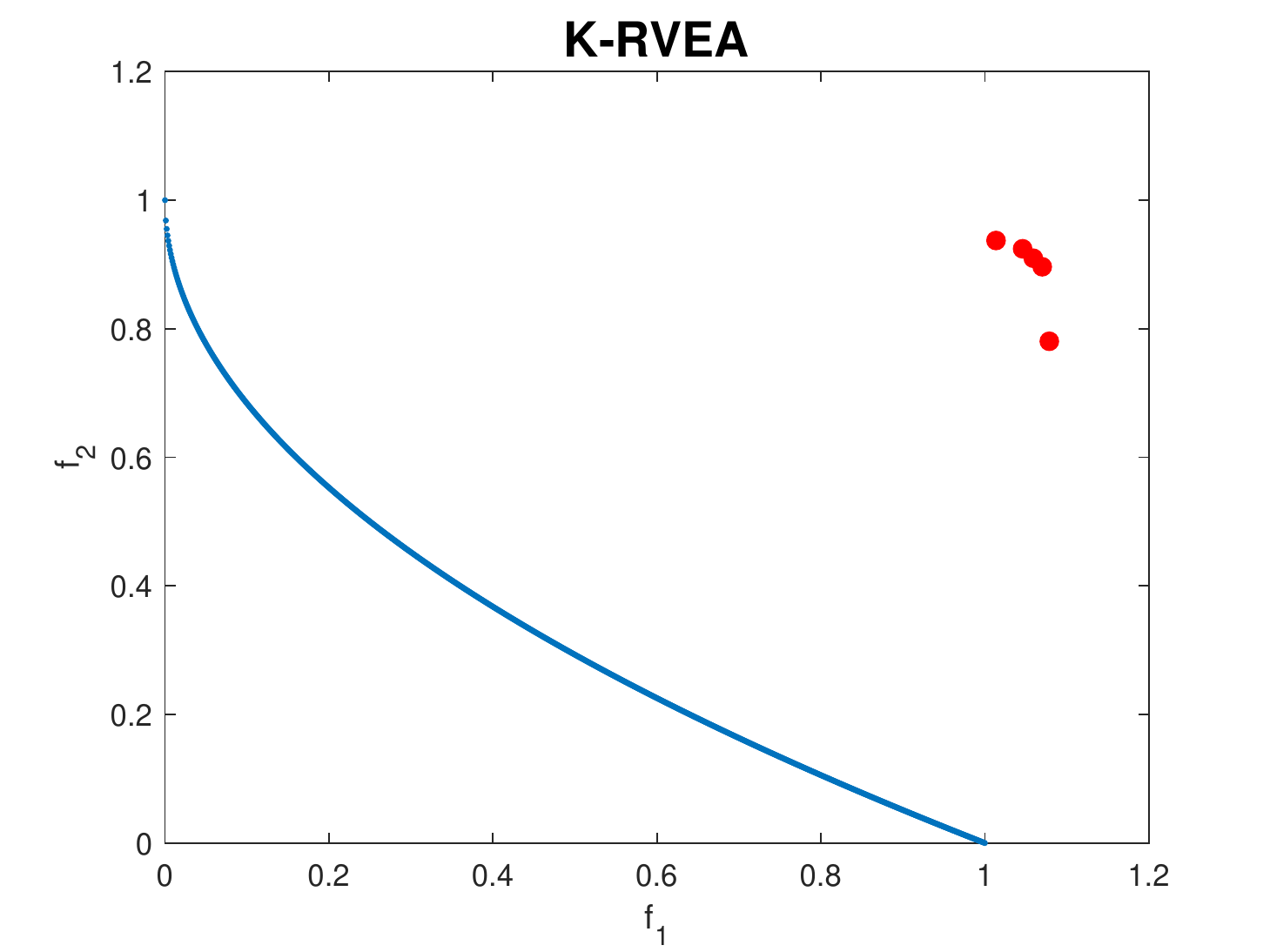}%
\hfil
}
\subfloat[]{\includegraphics[width=1.9in]{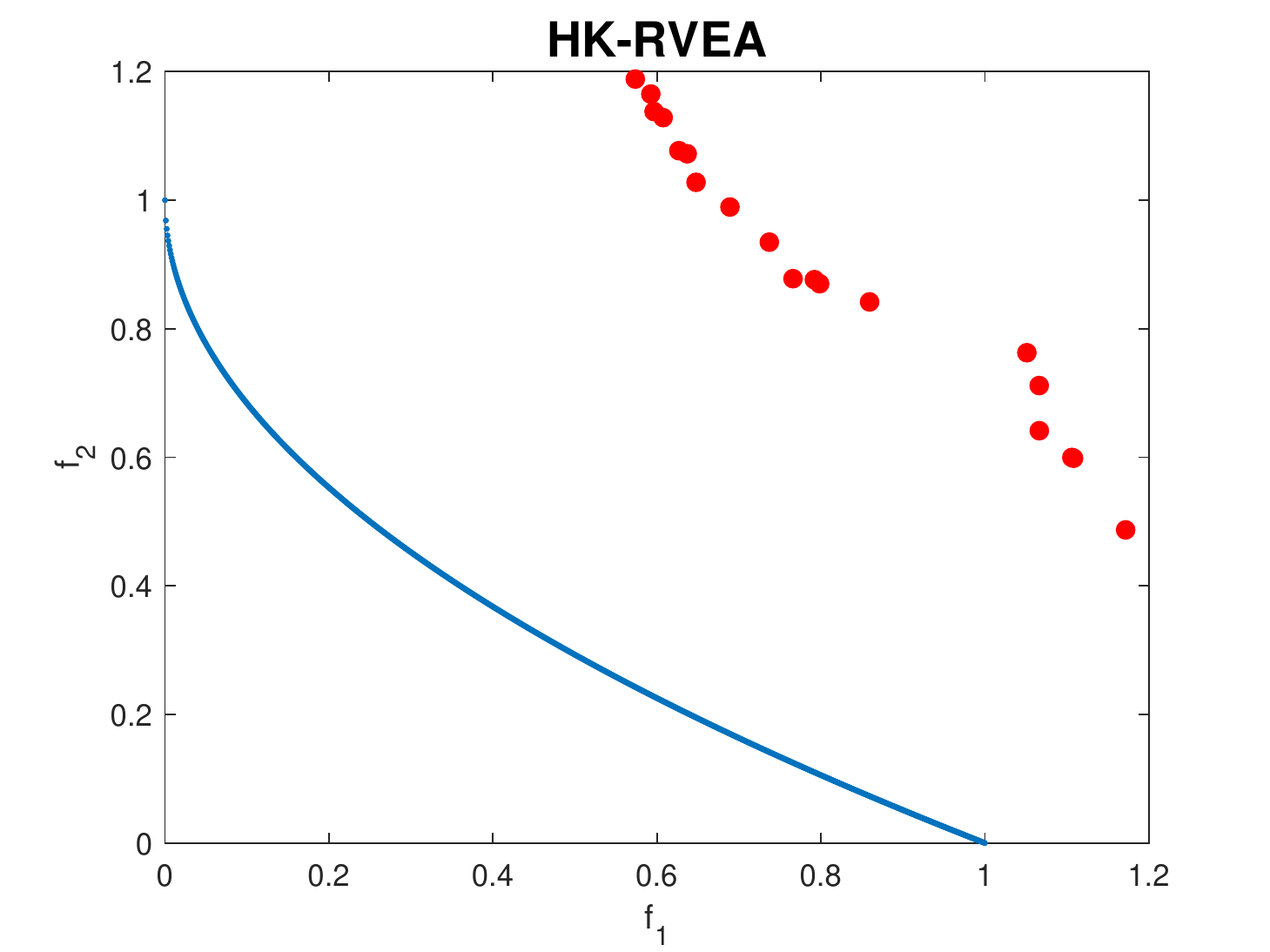}%
\hfil
}
\subfloat[]{\includegraphics[width=1.9in]{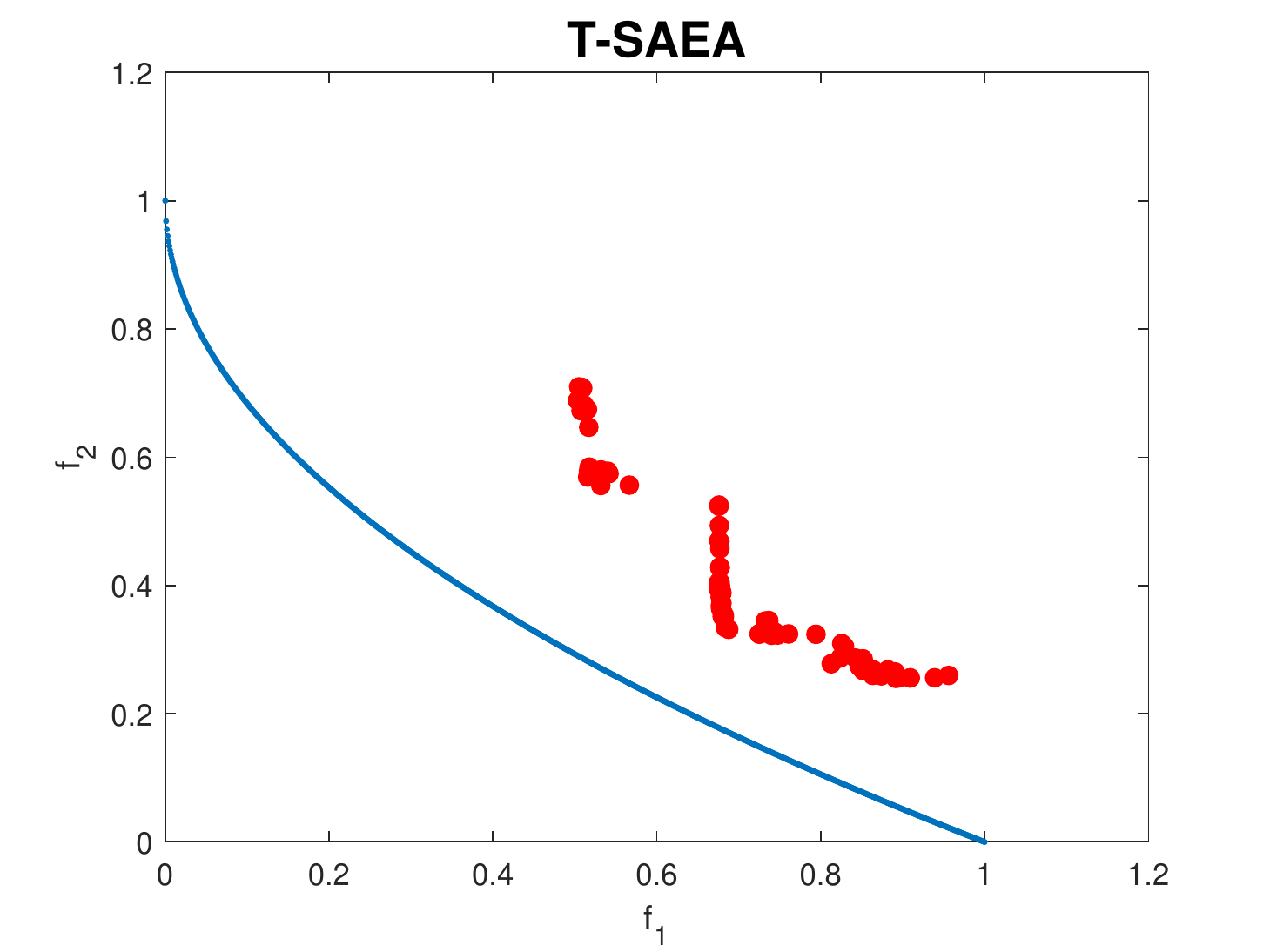}%
\hfil
}
\subfloat[]{\includegraphics[width=1.9in]{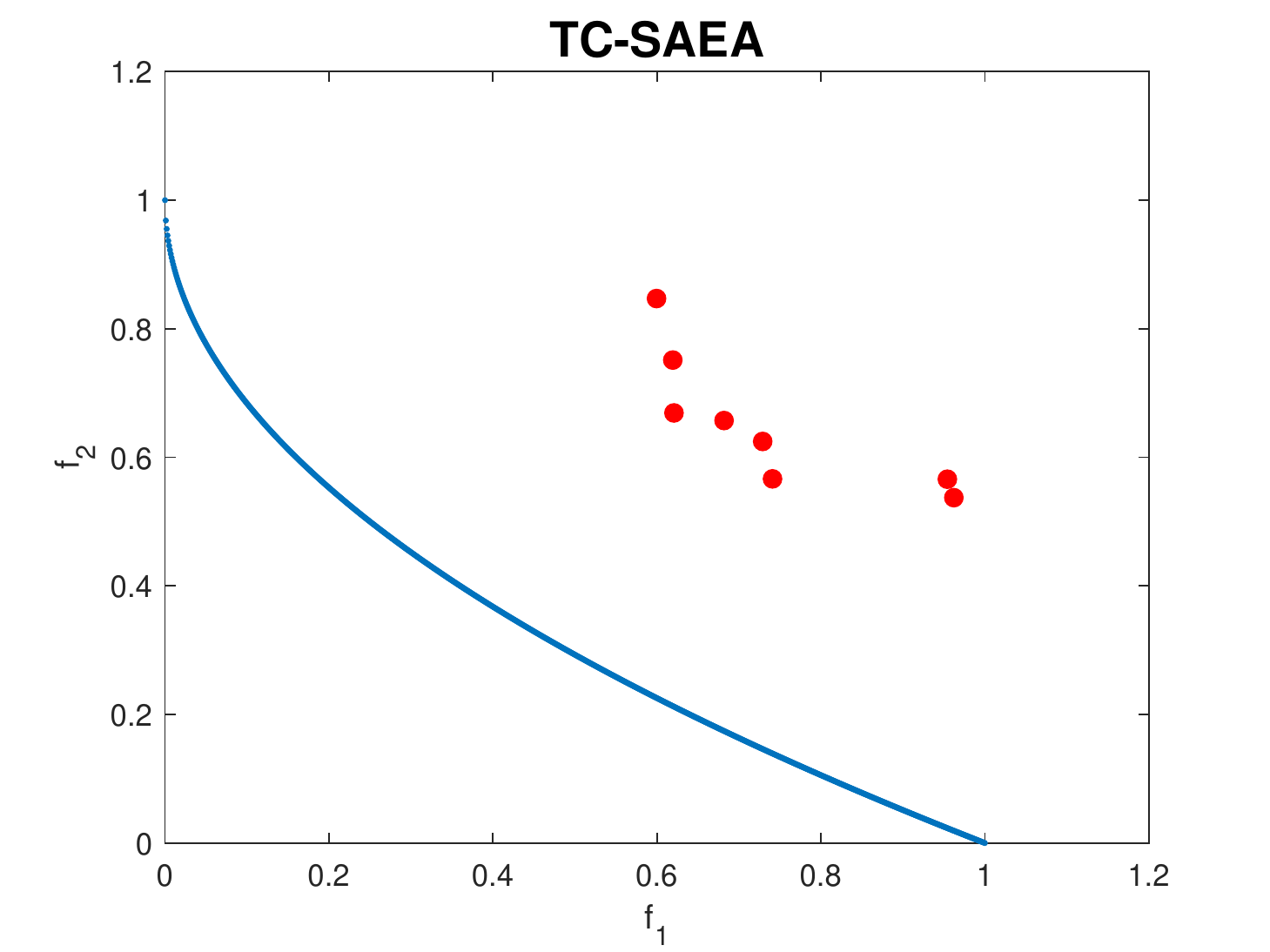}%
\hfil
}
}
\caption{The final non-dominated solutions obtained by the compared algorithms on the bi-objective UF3 in the run associated with the median IGD value.}
\label{Fig.6}
\end{figure*}

In this subsection, the performance of TC-SAEA on the above test instances is compared with the aforementioned algorithms in terms of convergence and diversity, indicated by IGD values. To investigate how different the delay length $\tau$ affects the algorithm's performance, the settings $\tau=5$ and $\tau=10$ are used, and the mean values and the standard deviations of IGD over 20 times are collected and presented in Tables \ref{Tab.1} and \ref{Tab.2}, where the best result of each benchmark function is highlighted. To further illustrate the advantage of the proposed algorithm, the non-dominated solution set obtained by each algorithm (in the run that achieved the medium performance out of 20 independent runs) on DTLZ2, DTLZ7 and UF3 are visualized in Figs. \ref{Fig.4}-\ref{Fig.6}.

According to the results in Table \ref{Tab.1}, we can observe that TC-SAEA outperforms the compared algorithms on all test instances except on DTLZ4, UF3 and UF7, demonstrating the effectiveness of the suggested strategies for solving bi-objective optimization problems with a delayed objective. The reason for TC-SAEA being outperformed by HK-RVEA on DTLZ4 may be that the density of the points on the true PF of DTLZ4 is strongly biased, and as a result, the synthetic values $\mathbf{Y}_{s}^{'a}$ in TC-SAEA are too far away from the true objective values of $f_{s}$. Consequently, the knowledge learned from the auxiliary data set $D_{a}={(\mathbf{X}^{a}, \mathbf{Y}_{s}^{'a})}$ is very limited and the risk of misleading the optimization process will be increased. It is important to observe that TC-SAEA is successful in accelerating convergence and achieving evenly distributed non-dominated solutions on the rest problems of the DTLZ test suite. The non-dominated solutions obtained by each algorithm on DTLZ2 and DTLZ7 are presented in Fig. \ref{Fig.4} and Fig. \ref{Fig.5}, respectively, confirming the effectiveness of the proposed instance transfer together with the co-surrogate. Similar conclusion can be drawn from the results on UF1-UF7. Figs. \ref{Fig.6} shows the performance of each algorithm on UF3, indicating that TC-SAEA is more likely to converge to the true PF with the help of knowledge transfer.

\begin{figure}[ht]
\centering
\includegraphics[width=0.5\columnwidth]{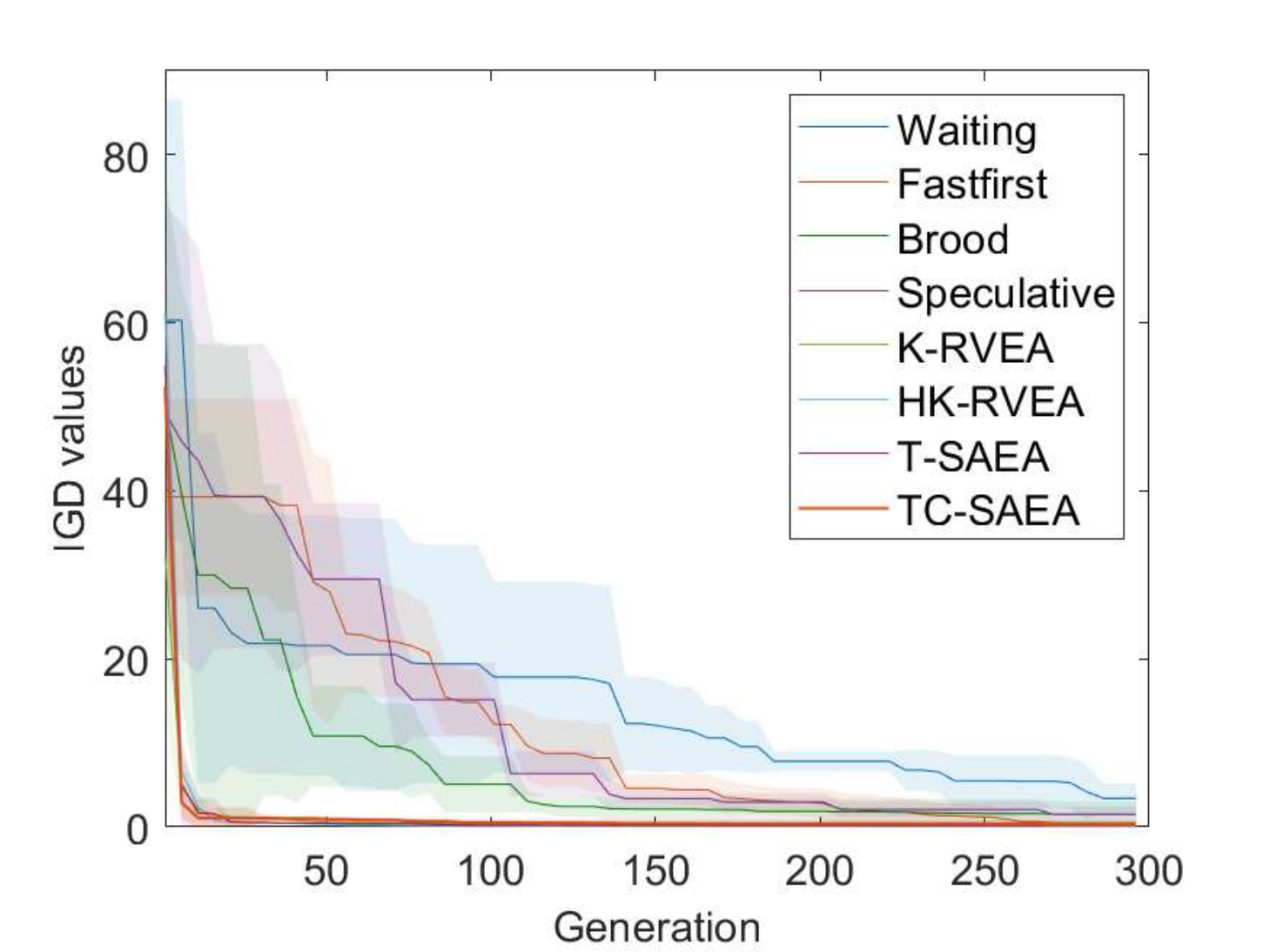}
\caption{IGD values obtained on DTLZ1a by each algorithm over the generations.}
\label{Fig.DTLZ1a1000}
\end{figure}

Our observations in terms of the three delay-handling schemes (i.e. \emph{Waiting}, \emph{Fast-first} and \emph{Interleaving schemes}) agree with the point made in \citep{allmendinger2015multiobjective}: \emph{Interleaving schemes} are generally the best. We note that the surrogate based delay-handing methods, such as HK-RVEA and T-SAEA, generally show better performance than the non-surrogate based ones on the problems with a limited budget of FEs, which can be explained in the sense that the former can benefit from surrogate models, while the latter suffers a slow convergence. It is interesting to compare K-RVEA with the surrogate based methods as K-RVEA is a surrogate based \emph{Waiting} method and HK-RVEA, T-SAEA and the proposed method are surrogate based \emph{Interleaving} methods. We can observe that by learning from the additional knowledge obtained on $f_f$, one can achieve a better trade-off between the convergence and diversity on bi-objective problems with heterogeneous objectives. To further demonstrate how the learning process proposed in our work helps the optimization, we run each algorithm on DTLZ1a with $FE_{s}^{max}=1000$, and the IGD values of the solution set over the generations obtained by each algorithm is plotted in Fig. \ref {Fig.DTLZ1a1000}, indicating the fast convergence of TC-SAEA.

To assess the impact of the delay length on the performance, Table \ref{Tab.2} presents the results obtained by each algorithm when $\tau=10$. Note that the relationship between the objective functions or the fitness landscape is unknown in our test problems. Therefore, it is unsurprising to observe that the performance on some test problems is degraded due to the increasing search bias resulting from the much more intensive search on the fast objective, even though the evaluation budget for the non-delayed objective is increased. For example, the performance of HK-RVEA on DTLZ1, DTLZ1a, DTLZ2, DTLZ3a and DTLZ6 is impacted negatively by a larger $\tau$. It is also noteworthy that TC-SAEA is still the most promising delay-handling method in comparison with the other algorithms in terms of diversity and convergence, which is consistent with the observations in Table \ref{Tab.1}.

\textcolor[rgb]{1,0,0}{We also compare TC-SAEA with Tr-SAEA \citep{wang2021transfer}, a recently reported algorithm for solving bi-objective problems with non-uniform evaluation times. The performance of K-RVEA on the same test instances but in the undelayed environment where all objectives are fast objectives and have the same time complexity is also included for reference to the best achievable results. The results in terms of IGD values on test instance with $\tau=5$ and $\tau=10$ are presented in Table \ref{Tab.add}. Although TC-SAEA and Tr-SAEA show similar performance on most test instances, it should be noted that TC-SAEA outperforms Tr-SAEA on DTLZ1 and DTLZ3. As suggested in \citep{deb2002scalable}, DTLZ1 and DTLZ3 can be adopted to investigate an MOEA’s ability to converge to the Pareto front. Hence, these results confirm the desirable strong convergence capability of TC-SAEA, implying potential advantages of direct knowledge transfer between different objectives. }

\begin{table}[!htbp]
\center
\caption{Statistical results of the IGD values obtained by Tr-SAEA, TC-SAEA and Undelayed algorithm with $FE_{s}^{max}=200$ and different $\tau$}
\label{Tab.add}
\setlength{\tabcolsep}{0.5mm}{
\begin{tabular}{l|ccc|cc|cc|ccc|cc|cc}
\hline

\multicolumn{1}{c|}{\multirow{2}{*}{Problem}} & \multicolumn{3}{c|}{Tr-SAEA}     & \multicolumn{2}{c|}{TC-SAEA} & \multicolumn{2}{c|}{Undelayed} & \multicolumn{3}{c|}{Tr-SAEA}     & \multicolumn{2}{c|}{TC-SAEA} & \multicolumn{2}{c|}{Undelayed} \\
\multicolumn{1}{c|}{}                         & mean          &           & std  & mean              & std      & mean           & std           & mean          &           & std  & mean              & std      & mean           & std           \\ \hline
\multicolumn{1}{c|}{} & \multicolumn{7}{c|}{$\tau=5$} & \multicolumn{7}{c}{$\tau=10$} \\  \hline
DTLZ1                                         & 20.7          & $\approx$ & 5.38 & \textbf{20.1}     & 8.16     & 15.8           & 5.71          & 29.5          & +         & 18.3 & \textbf{16.8}     & 13.9     & 12.2           & 5.52          \\
DTLZ1a                                        & \textbf{0.21} &  $\approx$   & 0.07 & 0.36              & 0.04     & 0.32           & 0.08          & \textbf{0.25} & --        & 0.05 & 0.60              & 0.40     & 0.23           & 0.04          \\
DTLZ2                                         & 0.03          & $\approx$ & 0.01 & \textbf{0.02}     & 0.00     & 0.02           & 0.01          & \textbf{0.03} & $\approx$ & 0.01 & \textbf{0.03}     & 0.02     & 0.02           & 0.00          \\
DTLZ3                                         & 327           & +         & 82.1 & \textbf{132}      & 79.28    & 168            & 61.9          & 348           & +         & 41.6 & \textbf{137}      & 77.0     & 119            & 24.5          \\
DTLZ3a                                        & 3.39          & $\approx$ & 1.87 & \textbf{2.30}     & 0.66     & 13.9           & 2.32          & 4.84          & +         & 2.97 & \textbf{0.43}     & 0.21     & 9.00           & 1.64          \\
DTLZ4                                         & \textbf{0.16} & --        & 0.07 & 0.44              & 0.13     & 0.05           & 0.07          & \textbf{0.19} & --        & 0.11 & 0.48              & 0.17     & 0.02           & 0.00          \\
DTLZ5                                         & \textbf{0.03} & $\approx$ & 0.03 & \textbf{0.03}     & 0.00     & 0.02           & 0.00          & 0.04          & $\approx$ & 0.01 & \textbf{0.03}     & 0.01     & 0.02           & 0.00          \\
DTLZ6                                         & \textbf{0.72} & --        & 0.09 & 2.62              & 1.95     & 2.52           & 0.04          & \textbf{0.74} & --        & 0.88 & 3.31              & 0.65     & 1.80           & 0.43          \\
DTLZ7                                         & \textbf{0.03} & $\approx$ & 0.01 & 0.05              & 0.08     & 0.04           & 0.00          & \textbf{0.02} & $\approx$ & 0.01 & 0.08              & 0.21     & 0.05           & 0.01          \\
UF1                                           & \textbf{0.19} & $\approx$ & 0.01 & \textbf{0.19}     & 0.02     & 0.15           & 0.04          & 0.39          & +         & 0.10 & \textbf{0.15}     & 0.02     & 0.15           & 0.02          \\
UF2                                           & \textbf{0.12} & $\approx$ & 0.01 & 0.13              & 0.02     & 0.09           & 0.02          & 0.14          & $\approx$ & 0.01 & \textbf{0.12}     & 0.02     & 0.08           & 0.01          \\
UF3                                           & 0.49          & $\approx$ & 0.01 & \textbf{0.42}     & 0.03     & 0.44           & 0.02          & \textbf{0.50} & $\approx$ & 0.03 & 0.51              & 0.06     & 0.42           & 0.02          \\
UF4                                           & 0.22          & $\approx$ & 0.00 & \textbf{0.19}     & 0.01     & 0.12           & 0.01          & \textbf{0.22} & $\approx$ & 0.00 & \textbf{0.22}     & 0.00     & 0.10           & 0.00          \\
UF5                                           & 2.43          & $\approx$ & 0.28 & \textbf{2.42}     & 0.38     & 1.55           & 0.31          & 2.81          & +         & 0.40 & \textbf{2.39}     & 0.20     & 1.22           & 0.23          \\
UF6                                           & 1.32          & +         & 0.39 & \textbf{0.81}     & 0.19     & 0.58           & 0.13          & 1.14          & +         & 0.27 & \textbf{0.69}     & 0.12     & 0.52           & 0.05          \\
UF7                                           & \textbf{0.32} & $\approx$ & 0.11 & 0.33              & 0.05     & 0.32           & 0.13          & 0.36          & +         & 0.07 & \textbf{0.27}     & 0.13     & 0.28           & 0.20          \\ \hline
\end{tabular}}
\end{table}
\subsection{Ablation Studies \label{Ablation Studies}}

\begin{table}[!htbp]
\center
\caption{Statistical results of the IGD values obtained by TC-SAEAp, NT-SAEA, NS-SAEA and TC-SAEA with $FE_{s}^{max}=200$ and $\tau=5$}
\label{Tab.3}
\setlength{\tabcolsep}{0.5mm}{
\begin{tabular}{l|lcc|ccc|ccc|cc}
\hline
\multicolumn{1}{c|}{\multirow{2}{*}{Problem}} & \multicolumn{3}{c|}{TC-SAEAp} & \multicolumn{3}{c|}{NT-SAEA} & \multicolumn{3}{c|}{NS-SAEA} & \multicolumn{2}{c}{TC-SAEA} \\
\multicolumn{1}{c|}{} & \multicolumn{2}{c}{mean} & std & \multicolumn{2}{c}{mean} & std & \multicolumn{2}{c}{mean} & std & mean & std \\ \hline
DTLZ1 & 26.6 & $\approx$ & 10.5 & 46.3 & + & 12.9 & 30.0 & + & 12.7 & \textbf{20.1} & 8.16 \\
DTLZ1a & 19.9 & + & 0.19 & 0.87 & + & 0.14 & 1.35 & + & 0.35 & \textbf{0.36} & 0.04 \\
DTLZ2 & 0.02 & $\approx$ & 0.03 & 0.05 & + & 0.01 & 0.03 & $\approx$ & 0.01 & 0.02 & 0.00 \\
DTLZ3 & 467 & + & 72.9 & 322 & + & 140 & 241 & + & 163 & \textbf{132} & 79.28 \\
DTLZ3a & 155 & + & 5.28 & 10.1 & + & 15.0 & 2.59 & + & 0.89 & \textbf{2.30} & 0.66 \\
DTLZ4 & 0.55 & + & 0.12 & \textbf{0.29} & -- & 0.52 & 0.52 & + & 0.19 & 0.44 & 0.17 \\
DTLZ5 & \textbf{0.02} & -- & 0.03 & 0.09 & + & 0.03 & 0.07 & + & 0.01 & 0.03 & 0.00 \\
DTLZ6 & 3.76 & + & 0.54 & 6.60 & + & 7.18 & 3.04 & + & 2.36 & \textbf{2.62} & 1.95 \\
DTLZ7 & \textbf{0.02} & -- & 0.06 & 1.29 & + & 0.40 & 0.45 & + & 0.29 & 0.05 & 0.08 \\
UF1 & 0.79 & + & 0.03 & 1.35 & + & 0.03 & 0.24 & + & 0.03 & \textbf{0.19} & 0.02 \\
UF2 & 0.14 & $\approx$ & 0.03 & 0.61 & + & 0.05 & 0.20 & + & 0.06 & \textbf{0.13} & 0.01 \\
UF3 & 0.92 & + & 0.06 & 0.55 & + & 0.01 & 0.52 & + & 0.02 & \textbf{0.42} & 0.03 \\
UF4 & 0.22 & $\approx$ & 0.00 & 0.19 & $\approx$ & 0.01 & 0.22 & $\approx$ & 0.02 & 0.19 & 0.01 \\
UF5 & 3.24 & + & 0.44 & 4.32 & + & 1.07 & 2.88 & + & 0.61 & \textbf{2.42} & 0.38 \\
UF6 & 2.14 & + & 0.13 & 4.13 & + & 3.06 & 0.98 & $\approx$ & 0.12 & \textbf{0.81} & 0.19 \\
UF7 & 0.67 & + & 0.05 & 0.50 & + & 0.09 & 0.37 & + & 0.09 & \textbf{0.33} & 0.05 \\
\hline
\end{tabular}}
\end{table}

\begin{table}[!htbp]
\center
\caption{Statistical results of the IGD values obtained by TC-SAEAp, NT-SAEA, NS-SAEA and TC-SAEA with $FE_{s}^{max}=200$ and $\tau=10$}
\label{Tab.4}
\setlength{\tabcolsep}{0.5mm}{
\begin{tabular}{l|lcl|ccc|ccc|cc}
\hline
\multicolumn{1}{c|}{\multirow{2}{*}{Problem}} & \multicolumn{3}{c|}{TC-SAEAp} & \multicolumn{3}{c|}{NT-SAEA} & \multicolumn{3}{c|}{NS-SAEA} & \multicolumn{2}{c}{TC-SAEA} \\
\multicolumn{1}{c|}{} & \multicolumn{2}{c}{mean} & \multicolumn{1}{c|}{std} & \multicolumn{2}{c}{mean} & std & \multicolumn{2}{c}{mean} & std & mean & std \\ \hline
DTLZ1 & 34.1 & + & 29.9 & 50.7 & + & 20.4 & 27.9 & + & 18.60 & \textbf{16.8} & 13.91 \\
DTLZ1a & 1.27 & + & 1.26 & 0.75 & + & 0.10 & \textbf{0.48} & -- & 0.16 & 0.60 & 0.40 \\
DTLZ2 & 0.03 & $\approx$ & 0.02 & 0.03 & $\approx$ & 0.00 & 0.11 & + & 0.09 & 0.03 & 0.02 \\
DTLZ3 & 376 & + & 141 & 382 & + & 137 & 336 & + & 85.0 & \textbf{137} & 77.0 \\
DTLZ3a & 314.8 & + & 546 & 11.6 & + & 12.6 & 3.65 & + & 4.08 & \textbf{0.43} & 0.21 \\
DTLZ4 & 0.92 & + & 0.27 & 0.67 & + & 0.30 & 0.54 & $\approx$ & 0.16 & \textbf{0.48} & 0.17 \\
DTLZ5 & \textbf{0.02} & -- & 0.01 & 0.03 & $\approx$ & 0.01 & 0.05 & + & 0.05 & 0.03 & 0.03 \\
DTLZ6 & 4.41 & + & 0.48 & 6.46 & + & 1.18 & \textbf{3.50} & $\approx$ & 0.62 & 3.51 & 0.65 \\
DTLZ7 & 0.16 & + & 0.00 & 0.13 & + & 0.10 & 0.45 & + & 0.00 & \textbf{0.08} & 0.29 \\
UF1 & 0.39 & + & 0.12 & 0.44 & + & 0.26 & 0.24 & + & 0.03 & \textbf{0.15} & 0.02 \\
UF2 & 0.15 & + & 0.03 & 0.27 & + & 0.01 & 0.23 & + & 0.07 & \textbf{0.12} & 0.02 \\
UF3 & 0.68 & + & 0.27 & 0.59 & + & 0.08 & 0.55 & + & 0.04 & \textbf{0.51} & 0.06 \\
UF4 & 0.22 & $\approx$ & 0.01 & \textbf{0.19} & -- & 0.01 & 0.23 & $\approx$ & 0.01 & 0.22 & 0.01 \\
UF5 & 2.76 & $\approx$ & 0.64 & 3.78 & + & 0.86 & 2.60 & $\approx$ & 0.48 & \textbf{2.39} & 0.28 \\
UF6 & 1.81 & + & 0.83 & 2.31 & + & 0.06 & 1.00 & + & 0.26 & \textbf{0.69} & 0.05 \\
UF7 & 0.63 & + & 0.12 & 0.64 & + & 0.13 & 0.35 & + & 0.02 & \textbf{0.27} & 0.13 \\
\hline
\end{tabular}}
\end{table}
To effectively address the optimization problems with delayed objectives, we propose an instance-based transfer learning scheme in the framework of a GP-based SAEA, hoping to transfer the knowledge readily available from the fast objective function for the optimization of the slow one. The proposed TC-SAEA is composed of two main components, a co-surrogate model to capture the hidden correlation between the objectives and a transferable instance selection strategy to identify the useful knowledge. In the following, we compare TC-SAEA with its three variants to further investigate the performance impact of the suggested strategies.
\begin{itemize}
\item A GP-based SAEA without the transfer learning scheme (NT-SAEA): In order to demonstrate the effectiveness of the transfer learning scheme, an SAEA without the TL scheme is introduced. In NT-SAEA, the GP model is built separately for each objective with different available training data sets. Due to the different evaluation times, the number of new samples for updating each surrogate is also different.
\item A GP-based SAEA with a polynomial regression as the co-surrogate (TC-SAEAp): Instead of the GP, a polynomial regression is used as the co-surrogate to model the correlation between the objectives. The proposed TL scheme is adopted in TC-SAEAp.
\item A GP-based SAEA without the transferable instance selection (NS-SAEA): In this variant, only the co-surrogate component in the proposed TL scheme is adopted in NS-SAEA. Due to the lack of selection, all the auxiliary data $D_{a}={(\mathbf{X}^{a}, \mathbf{Y}_{s}^{'a})}$ will be added to the training data set of the slow objective.
\end{itemize}

Firstly, it is observed that TC-SAEA achieves the better performance than TC-SAEAp in terms of convergence and diversity on both test instances with $\tau=5$ and $\tau=10$. Recall that the difference between TC-SAEA and TC-SAEAp lies in their co-surrogate model; therefore, this observation confirms that the GP co-surrogate model can capture the hidden correlation between objectives more accurately and effectively than the polynomial regression model. Secondly, it is clear to see from Tables \ref{Tab.3} and \ref{Tab.4} that TC-SAEA has superior performance to NT-SAEA on most test instances, while NT-SAEA yields the better results on DTLZ4 when $\tau=5$ and on UF4 when $\tau=10$. Compared with NT-SAEA, TC-SAEA is able to learn some useful knowledge from the auxiliary data set, when the observed data for $f_{s}$ is insufficient to train a good model. The comparison between NT-SAEA and TC-SAEA confirms that the performance of the surrogate model $GP_{s}$ for $f_{s}$ can benefit from the transferable data set $D_{t}$. Thirdly, it should be pointed out that NS-SAEA can achieve good performance on DTLZ1a and DTLZ6 only when $\tau=10$, which means that transferable instance selection is helpful in most cases. Despite the use of transfer learning in NS-SAEA, it fails to converge towards a set of acceptable solutions. This observation clearly indicates the benefit of the proposed transferable instance selection. Note that the effectiveness of the transfer learning scheme heavily relies on the quality of the auxiliary data $\mathbf{Y}_{s}^{'a}$ generated by transfer learning. Compared with the carefully selected transferable data, applying the auxiliary data directly to train $GP_{s}$ without filtering out unreliable synthetic data may be more likely to lead to negative transfer. TC-SAEA adopts an instance selection method according to the uncertainty level provided by $GP_{s}$ in order to mitigate possible negative transfer, which is demonstrated to be effective by the comparative results between TC-SAEA and NS-SAEA. In summary, it is evident that TC-SAEA is better suited for solving bi-objective optimization problems for objectives with distinct evaluation times in comparison with NT-SAEA and NS-SAEA.

\subsection{Impact of the Correlation between Objectives}
\begin{table*}[!htb]
\center
\caption{The IGD values obtained by Waiting, Fast-first, Brood interleaving, Speculative interleaving, K-RVEA, HK-RVEA, T-SAEA, and TC-SAEA on the cm-OneMax problem with $FE_{s}^{max}=200$}
\label{Tab.5}
\setlength{\tabcolsep}{0.15mm}{
\begin{tabular}{l|l|lll|lll|lll|lll|lll|lll|lll|ll}
\hline
\multicolumn{1}{c|}{\multirow{2}{*}{Problem}} & \multicolumn{1}{c|}{\multirow{2}{*}{$\tau$}} & \multicolumn{3}{c|}{Waiting}                         & \multicolumn{3}{c|}{Fast-first}                      & \multicolumn{3}{c|}{BI}                           & \multicolumn{3}{c|}{SI}      & \multicolumn{3}{c|}{K-RVEA}      & \multicolumn{3}{c|}{HK-RVEA}                         & \multicolumn{3}{c|}{T-SAEA}                          & \multicolumn{2}{c}{TC-SAEA}                        \\
\multicolumn{1}{c|}{}                           & \multicolumn{1}{c|}{}                        & \multicolumn{2}{c}{mean} & \multicolumn{1}{c|}{std} & \multicolumn{2}{c}{mean} & \multicolumn{1}{c|}{std} & \multicolumn{2}{c}{mean} & \multicolumn{1}{c|}{std} & \multicolumn{2}{l}{mean} & std       & \multicolumn{2}{l}{mean} & std  & \multicolumn{2}{c}{mean} & \multicolumn{1}{c|}{std} & \multicolumn{2}{c}{mean} & \multicolumn{1}{c|}{std} & \multicolumn{1}{c}{mean} & \multicolumn{1}{c}{std} \\ \hline
\multirow{2}{*}{corr=-1}                        & $\tau$=5                                     & 0.41          & +         & 0.12                     & 2.26          & +         & 0.07                     & 0.52          & +         & 0.09                     & 0.45          & +         & 0.10      & 0.26          & +         & 0.11 & 0.20          & +         & 0.09                     & 0.18          & +         & 0.06                     & \textbf{0.13}            & 0.04                     \\
                                                & $\tau$=10                                    & 0.41          & +         & 0.12                     & 2.65          & +         & 0.15                     & 0.51          & +         & 0.11                     & 0.42          & +         & 0.07      & 0.26          & +         & 0.11 & 0.22          & +         & 0.07                     & 0.22          & +         & 0.08                     & \textbf{0.12}            & 0.05                     \\ \hline
\multirow{2}{*}{corr=-0.75}                     & $\tau$=5                                     & 0.90          & +         & 0.08                     & 1.82          & +         & 0.22                     & 1.13          & +         & 0.08                     & 1.09          & +         & 0.17      & 0.27          & +         & 0.03 & \textbf{0.14} & $\approx$ & 0.03                     & 0.46          & +         & 0.10                     & \textbf{0.14}            & 0.02                     \\
                                                & $\tau$=10                                    & 0.90          & +         & 0.08                     & 2.08          & +         & 0.25                     & 1.14          & +         & 0.07                     & 1.04          & +         & 0.09      & 0.27          & +         & 0.03 & \textbf{0.16} & $\approx$ & 0.03                     & 0.55          & +         & 0.15                     & \textbf{0.16}            & 0.05                     \\ \hline
\multirow{2}{*}{corr=-0.5}                      & $\tau$=5                                     & 0.97          & +         & 0.08                     & 1.82          & +         & 0.22                     & 1.11          & +         & 0.12                     & 1.02          & +         & 0.14      & 0.28          & +         & 0.05 & 0.13          & +         & 0.02                     & 0.43          & +         & 0.12                     & \textbf{0.10}            & 0.01                     \\
                                                & $\tau$=10                                    & 0.97          & +         & 0.08                     & 1.98          & +         & 0.28                     & 1.06          & +         & 0.07                     & 1.05          & +         & 0.14      & 0.28          & +         & 0.05 & 0.15          & +         & 0.04                     & 0.47          & +         & 0.10                     & \textbf{0.10}            & 0.04                     \\ \hline
\multirow{2}{*}{corr=-0.25}                     & $\tau$=5                                     & 1.20          & +         & 0.11                     & 1.83          & +         & 0.21                     & 1.32          & +         & 0.16                     & 1.35          & +         & 0.11      & 0.25          & +         & 0.03 & \textbf{0.13} & $\approx$ & 0.03                     & 0.42          & +         & 0.14                     & \textbf{0.13}            & 0.01                     \\
                                                & $\tau$=10                                    & 1.20          & +         & 0.11                     & 2.27          & +         & 0.21                     & 1.42          & +         & 0.10                     & 1.30          & +         & 0.13      & 0.25          & +         & 0.03 & \textbf{0.14} & $\approx$ & 0.03                     & 0.58          & +         & 0.20                     & \textbf{0.14}            & 0.02                     \\ \hline
\multirow{2}{*}{corr=0}                         & $\tau$=5                                     & 1.50          & +         & 0.21                     & 2.02          & +         & 0.30                     & 1.70          & +         & 0.18                     & 1.57          & +         & 0.16      & 0.21          & +         & 0.06 & \textbf{0.11} & $\approx$ & 0.02                     & 0.38          & +         & 0.09                     & \textbf{0.11}            & 0.02                     \\
                                                & $\tau$=10                                    & 1.50          & +         & 0.21                     & 2.35          & +         & 0.25                     & 1.60          & +         & 0.10                     & 1.50          & +         & 0.15      & 0.21          & +         & 0.06 & \textbf{0.11} & $\approx$ & 0.02                     & 0.51          & +         & 0.19                     & \textbf{0.11}            & 0.01                     \\ \hline
corr=0.25                                       & $\tau$=5                                     & 1.52          & +         & 0.15                     & 2.06          & +         & 0.22                     & 1.70          & +         & 0.12                     & 1.47          & +         & 0.13      & 0.19          & +         & 0.03 & \textbf{0.12} & $\approx$ & 0.01                     & 0.60          & +         & 0.12                     & \textbf{0.12}            & 0.01                     \\
                                                & $\tau$=10                                    & 1.52          & +         & 0.15                     & 2.28          & +         & 0.24                     & 1.71          & +         & 0.22                     & 1.53          & +         & 0.13      & 0.19          & +         & 0.03 & \textbf{0.12} & $\approx$ & 0.01                     & 0.62          & +         & 0.21                     & \textbf{0.12}            & 0.01                     \\ \hline
\multirow{2}{*}{corr=0.5}                       & $\tau$=5                                     & 1.78          & +         & 0.18                     & 2.32          & +         & 0.31                     & 2.01          & +         & 0.24                     & 1.75          & +         & 0.28      & 0.16          & +         & 0.04 & 0.14          & +         & 0.02                     & 0.45          & +         & 0.20                     & \textbf{0.12}            & 0.01                     \\
                                                & $\tau$=10                                    & 1.78          & +         & 0.18                     & 2.27          & +         & 0.34                     & 1.92          & +         & 0.23                     & 1.78          & +         & 0.22      & 0.16          & +         & 0.04 & 0.13          & +         & 0.01                     & 0.54          & +         & 0.11                     & \textbf{0.11}            & 0.01                     \\ \hline
\multirow{2}{*}{corr=0.75}                      & $\tau$=5                                     & 1.96          & +         & 0.30                     & 2.61          & +         & 0.35                     & 2.33          & +         & 0.27                     & 2.28          & +         & 0.26      & 0.13          & +         & 0.03 & 0.13          & +         & 0.01                     & 0.40          & +         & 0.17                     & \textbf{0.12}            & 0.02                     \\
                                                & $\tau$=10                                    & 1.96          & +         & 0.30                     & 2.60          & +         & 0.34                     & 2.31          & +         & 0.25                     & 1.87          & +         & 0.18      & 0.13          & +         & 0.03 & 0.13          & +         & 0.02                     & 0.39          & +         & 0.11                     & \textbf{0.12}            & 0.01                     \\ \hline
\multirow{2}{*}{corr=1}                         & $\tau$=5                                     & 2.77          & +         & 0.56                     & 3.86          & +         & 0.33                     & 3.66          & +         & 0.45                     & 3.16          & +         & 0.34      & \textbf{0.00} & $\approx$ & 0.00 & \textbf{0.00} & $\approx$ & 0.00                     & \textbf{0.00} & $\approx$ & 0.00                     & \textbf{0.00}            & 0.00                     \\
                                                & $\tau$=10                                    & 2.77          & +         & 0.56                     & 3.81          & +         & 0.55                     & 3.24          & +         & 0.67                     & 3.27          & +         & 0.33      & \textbf{0.00} & $\approx$ & 0.00 & \textbf{0.00} & $\approx$ & 0.00                     & \textbf{0.00} & $\approx$ & 0.00                     & \textbf{0.00}            & 0.00                     \\ \cline{2-25}

 \hline
\end{tabular}}
\end{table*}
The rationale behind the proposed transfer learning scheme is that there will be a certain degree of correlation between the objectives of the Pareto optimal solutions, although such a correlation may be weak or not exist in most part of the search space. Here, we further investigate the impact of the correlation between objectives on the performance of TC-SAEA using the cm-OneMax test problem, since the strength of the correlation between its objectives is controllable. The IGD values of the obtained solutions by each algorithm under comparison on the cm-OneMax problem with $corr={-1, -0.75, -0.5, -0.25, 0, 0.25, 0.5, 0.75, 1}$ are presented in Table \ref{Tab.5}.

These results indicate that the proposed TC-SAEA exhibits the best performance on the cm-OneMax problem when there is a positive or negative correlation between the objectives, and HK-RVEA is the second best. This observation demonstrates the ability of the proposed TC-SAEA for transferring useful knowledge between the objectives. Interestingly, if we take into account the results obtained on the DTLZ and UF test suites, where the correlation between the objectives on the PF is defined by a function, we can conclude that TC-SAEA generally has achieved good performance, confirming its ability to utilize the knowledge obtained from $f_f$ to assist the optimization of $f_s$. The enhanced performance of TC-SAEA can be attributed to capability of the co-surrogate of capturing the relationship between the objectives and the instance selection strategy to reduce the negative transfer. Secondly, HK-RVEA has also obtained good performance on the cm-OneMax with $corr={-0.75, -0.25, 0, 0.25, 1}$, which is consistent with the results in \citep{chugh2018surrogate}. Thirdly, it is observed that \emph{Waiting} shows stable and good performance compared with the delay-handling methods without surrogates, particularly in case there is a negative correlation between the objectives. However, accounting for expensive MOPs with a limited amount of evaluation budgets, surrogate assisted methods, such as K-RVEA, HK-RVEA and TC-SAEA, are more promising compared with methods without surrogates.

\subsection{\textcolor[rgb]{1,0,0}{Effects of the Co-surrogate}}
\label{sec:co-surrogate}
\textcolor[rgb]{1,0,0}{To gain deeper insights into the relationship between the performance of TC-SAEA and the approximation quality of the co-surrogate $GP_c$, we examine the mean square error (MSE) of the estimated difference between $f_s$ and $f_f$ on the additional solutions $\mathbf{X}^a$ provided by $GP_c$. Specifically, we run TC-SAEA on each bi-objective DTLZ test instance and calculate the MSE of the estimation provided by the co-surrogate on $\mathbf{X}^a$ at each generation. Ten independent runs, each with a maximum of 200 expensive function evaluations, are performed and the mean and variance of the MSE of the co-surrogate are calculated and plotted in Fig. S1 in the Supplementary material. We also plot the achieved non-dominated solutions for each instance in the run associated with the median IGD value in Fig. S2. From these results, we can observe that the approximation error of the co-surrogate converges over the generations if the obtained non-dominated solutions are close to the Pareto front, such as on DTLZ2, DTLZ3a, DTLZ4, DTLZ5 and DTLZ7. By contrast, the approximation error of the co-surrogate strongly oscillates on DTLZ1, DTLZ2, and DTLZ6 because the obtained non-dominated solutions are still far away from the true Pareto front, where no clear relationship between $f_s$ and $f_f$ exists. These results agree well with our hypothesis discussed in Section \ref{sec:co-surr}.}

\subsection{Parameter Sensitivity Analysis}
The sensitivity analysis of the performance to the number of new samples $u$ for updating GPs in TC-SAEA is presented in the Supplementary material. Fig. S3 in the Supplementary material shows the boxplots in terms of IGD over different $u$ values on the DTLZ test suite. We can see that TC-SAEA with different $u$ exhibits distinct performance on each test instance, indicating that the best setting of $u$ may be problem-dependent. For example, TC-SAEA shows similar performance on DTLZ2 and DTLZ3a as $u$ changes , which is not the case on DTLZ3 and DTLZ4. Interestingly, setting too small or too large values to $u$, e.g., $u=1$ or $u=10$, will negatively impact the performance of TC-SAEA on most test instances. By contrast, TC-SAEA with $u=3, 5$ show competitive results. Therefore, $u=3$ is considered to be appropriate for our algorithm for handling bi-objective optimization problems with a delayed objective.

\section{Conclusion}
In this paper we focus on the bi-objective optimization problems where the evaluation of one objective takes longer time than another one, and propose a transfer learning scheme within a GP-assisted multi-objective evolutionary algorithm. The transfer learning scheme includes two key components, a co-surrogate model to indirectly relate the fast objective function to the slow objective function and a transferable instance selection to identify the useful knowledge in the auxiliary dataset. The co-surrogate model is used to generate synthetic data for the expensive objective function by transferring knowledge from additional function evaluations of the fast objective. To reduce the risk of negative transfer, the confidence bound of the surrogate for the slow objective function is used to filter out unreliable synthetic data.

The proposed algorithm is tested on sets of widely used benchmark problems for different delay lengths. Our experimental results demonstrate that the proposed algorithm achieves significantly better performance than the state-of-the-art delay-handling methods on most test instances studied in this work. Comparisons are also carried out to investigate the effectiveness of the individual mechanisms of the proposed algorithm and the empirical results equally confirm that both the co-surrogate and the transferable instance selection are indispensable for the good performance of the proposed algorithm.

Despite the encouraging results, research on MOPs with delayed objectives is still in its infancy stage and many challenges remain to be addressed in the future. For example, it will become more challenging to achieve efficient knowledge transfer when the number of objectives increases. In addition, it is of interest to investigate the most effective way of making use of the additional function evaluations of the cheap objective. Finally, an optimal resource allocation between fast and slow objective is conceivable to achieve the best performance for a given total computational budget.

\small
\bibliographystyle{apalike}
\bibliography{ecjsample}

\end{document}